\documentclass[11pt]{article}

\usepackage[a4paper,margin=1in]{geometry}
\usepackage{amsmath,amssymb,amsfonts}
\usepackage{booktabs}
\usepackage{array}
\usepackage{enumitem}

\usepackage{xurl}
\usepackage{hyperref}
\newcommand{\doi}[1]{\href{https://doi.org/#1}{\nolinkurl{doi:#1}}}
\usepackage[numbers,sort&compress]{natbib}

\usepackage{graphicx}
\usepackage{xcolor}
\usepackage{tikz}

\usetikzlibrary{
	positioning,
	arrows.meta,
	shapes.geometric,
	calc,
	fit,
	backgrounds
}

\usepackage{adjustbox}
\usepackage{placeins}
\usepackage{makecell}

\title{Verification of Adaptive Agentic Controllers through Finite Rule Revision}

\author{
	Roberto Garrone\\
	Open University of Cyprus\\
	\texttt{roberto.garrone@st.ouc.ac.cy}
}

\date{}

\begin{document}
	
	\maketitle
	
	\begin{abstract}
		Industrial agentic AI systems increasingly exhibit a gap between prototype capability and production deployment. In particular, adaptive agents may generate plausible outputs while remaining difficult to verify under non-determinism, confidentiality constraints, limited context, and weak observability. This paper formulates a bounded verification protocol for adaptive agentic controllers represented by finite symbolic rules, explicit diagnostic predicates, explanation logs, and held-out re-evaluation. The central research question is: when an adaptive agentic controller is represented through finite rules, explicit diagnostic predicates, explanation logs, and held-out re-evaluation, which classes of controller failure can be detected, locally repaired, or rejected without relying on unrestricted human-in-the-loop judgment? The proposed framework treats the controller as a finite revisable object. Diagnostic failures are mapped to predefined rule-level edits, including rule addition, rule deletion, and priority revision. Repaired controllers are then evaluated on held-out simulation seeds or cloned initial states. Experiments in a stylized financially constrained inventory-control benchmark show three outcomes: resource-induced failures that remain non-repairable by one rule edit, partial repairs that are rejected because they violate thresholds or guardrails, and a local one-step repair of an order-volatility failure induced by removing a smoothing rule. The contribution is methodological and provides a simulation-compatible procedure for testing whether specific controller-level failures can be made observable, explainable, locally revisable, and empirically re-tested under controlled conditions.
	\end{abstract}
	
	
\section{Introduction}

Agentic AI systems are increasingly used to execute multi-step tasks, coordinate tool use, retrieve contextual information, and adapt their actions to changing operational conditions. These capabilities make them relevant for enterprise automation, decision support, and workflow coordination. However, the transition from prototype demonstrations to production deployment remains limited by a verification problem. Recent empirical evidence on industrial agentic AI adoption identifies a capability-deployment verification gap: organizations may demonstrate higher-level agentic capabilities experimentally while being unable to integrate them into production workflows because adequate output-verification mechanisms are absent \citep{industrialgap2026}.

In enterprise settings, a common architectural response is to prevent the agent from becoming the system of record. The agent may interpret tasks, retrieve approved context, select allowed tools, and recommend actions, but execution should remain mediated by governed APIs, workflow engines, RPA layers, enterprise systems, and approval gates. This design principle is necessary because it separates reasoning and coordination from authoritative execution. It is not sufficient, however, because it does not by itself verify the behavior of the agentic controller. Even when execution is routed through governed systems, the controller may still select inappropriate actions, overreact to noisy signals, fail to respect diagnostic constraints, or produce decisions that cannot be explained and re-tested.

\paragraph{Deployment verification gap.}
In this paper, the deployment verification gap denotes the following technical condition. Let $A$ be an adaptive agentic controller that produces actions $a_t$ from observed states $s_t$ under stochastic or context-dependent conditions. A prototype demonstration establishes only that there exists at least one observed execution trace
\[
\tau = (s_0,a_0,\ldots,s_T)
\]
that satisfies an informal task objective. Production deployment, by contrast, requires evidence that a class of traces generated under admissible perturbations satisfies declared operational properties:
\[
\Pr_{\omega \sim \Omega}
\left[
\tau_A(\omega) \models \varphi
\right]
\geq 1-\alpha,
\]
where $\Omega$ is the evaluation distribution, $\varphi$ is a finite set of diagnostic and guardrail properties, and $\alpha$ is the tolerated failure probability. The verification gap arises when prototype capability is observed, but the mapping from controller logic to admissible behavior cannot be inspected, bounded, revised, or re-tested against $\varphi$. This formalization narrows the industrial verification problem identified by \citet{industrialgap2026} to the controller level.

\begin{figure}[h]
	\centering
	\resizebox{\textwidth}{!}{
		\begin{tikzpicture}[
			node distance=1.3cm,
			every node/.style={font=\small},
			box/.style={
				rectangle,
				rounded corners,
				draw=black,
				thick,
				align=center,
				minimum width=3.4cm,
				minimum height=1.0cm
			},
			layer/.style={
				rectangle,
				rounded corners,
				draw=black,
				thick,
				align=center,
				minimum width=15.2cm,
				minimum height=1.1cm,
				fill=gray!10
			},
			highlight/.style={
				rectangle,
				rounded corners,
				draw=black,
				ultra thick,
				align=center,
				minimum width=4.6cm,
				minimum height=1.25cm,
				fill=yellow!25
			},
			gapbox/.style={
				rectangle,
				rounded corners,
				draw=red!70!black,
				thick,
				align=center,
				fill=red!8,
				minimum width=6.6cm,
				minimum height=0.9cm
			},
			arrow/.style={->, thick},
			dashedarrow/.style={->, thick, dashed}
			]
			
			\node[layer] (strategy) {
				\textbf{Platform-Agnostic Strategy Layer}\\
				Business process decomposition $\;+\;$ agentic use-case design $\;+\;$ risk appetite $\;+\;$ verification requirements
			};
			
			\node[box, below=1.2cm of strategy, xshift=-6.0cm] (orch) {
				\textbf{Agent}\\
				\textbf{Orchestration}\\
				planning, routing,\\
				task decomposition
			};
			
			\node[box, right=0.55cm of orch] (data) {
				\textbf{Trusted Data}\\
				\textbf{Layer}\\
				RAG, knowledge base,\\
				semantic layer
			};
			
			\node[box, right=0.55cm of data] (tools) {
				\textbf{Governed}\\
				\textbf{Tools/APIs}\\
				ERP, CRM, workflows,\\
				models, services
			};
			
			\node[box, right=0.55cm of tools] (human) {
				\textbf{Human}\\
				\textbf{Approval}\\
				review, escalation,\\
				sign-off
			};
			
			\node[box, right=0.55cm of human] (obs) {
				\textbf{Observability}\\
				\textbf{\& Evaluation}\\
				logs, tests, metrics,\\
				monitoring
			};
			
			\node[highlight, below=3cm of obs, xshift=-1.15cm] (article) {
				\textbf{Controller-level verification}\\
				Finite rules, diagnostic predicates,\\
				explanation logs, held-out tests,\\
				bounded rule revision
			};
			
			\node[layer, below=1cm of article, xshift=-3cm] (implementation) {
				\textbf{Client-Stack-Specific Implementation Layer}\\
				Microsoft / AWS / Google / Salesforce / ServiceNow / Databricks / SAP / custom APIs / internal data platforms
			};
			
			\draw[arrow] (strategy.south) -- ++(0,-0.45) -| (orch.north);
			\draw[arrow] (strategy.south) -- ++(0,-0.45) -| (data.north);
			\draw[arrow] (strategy.south) -- ++(0,-0.45) -| (tools.north);
			\draw[arrow] (strategy.south) -- ++(0,-0.45) -| (human.north);
			\draw[arrow] (strategy.south) -- ++(0,-0.45) -| (obs.north);
			
			\draw[arrow] (orch.east) -- (data.west);
			\draw[arrow] (data.east) -- (tools.west);
			\draw[arrow] (tools.east) -- (human.west);
			\draw[arrow] (human.east) -- (obs.west);
			
			\draw[arrow] (obs.south) -- (article.north);
			\draw[dashedarrow] (article.west) -| (human.south);
			\draw[dashedarrow] (article.north west) -| (orch.south);
			\draw[dashedarrow] (article.south) -- (implementation.north);
			
			\node[gapbox, below=1cm of tools, xshift=-0.2cm] (gap) {
				\textbf{Deployment Verification Gap}\\
				Prototype capability exists, but production approval lacks sufficient evidence
			};
			
			\draw[dashedarrow] (gap.north west) -- (orch.south east);
			\draw[dashedarrow] (gap.north) -- (tools.south);
			\draw[dashedarrow] (gap.north east) -- (human.south west);
			\draw[dashedarrow] (gap.east) -- (obs.south west);
			
		\end{tikzpicture}
	}
	\caption{Placement of the controller-level verification problem within a generic enterprise architecture for agentic AI. The paper does not address the full deployment stack. It focuses on the verification layer that connects orchestration, observability, and approval.}
	\label{fig:bain_agentic_architecture}
\end{figure}
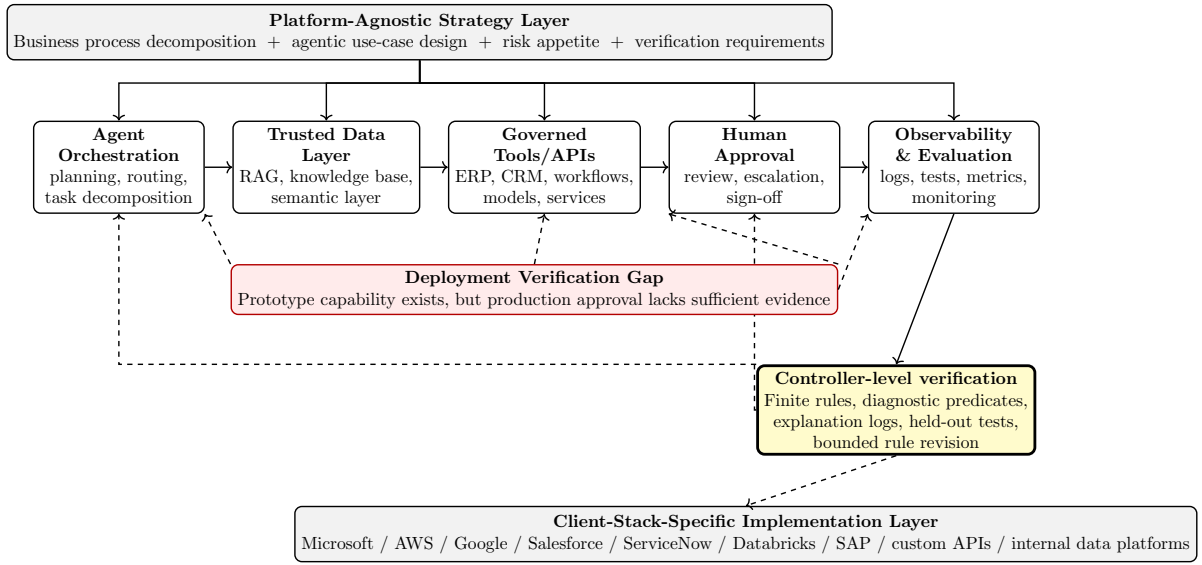

Figure~\ref{fig:bain_agentic_architecture} positions this problem within a generic enterprise architecture for agentic automation. The figure is not intended to define a complete deployment architecture. Its purpose is narrower: it shows that the paper addresses the controller-level verification problem located between orchestration, observability, and approval. Trusted data engineering, API governance, cybersecurity, legal review, and production integration remain outside the scope of the proposed method.

This paper studies one restricted part of the deployment verification problem: controller-level verification for adaptive agentic systems. The focus is not on open-ended conversational agents, unrestricted autonomous systems, or full enterprise deployment. Instead, the paper considers adaptive controllers whose decisions can be represented through finite rules, diagnostic predicates, explanation logs, and controlled simulation experiments. This restricted formulation builds on prior work on adaptive, explainable, and contestable agent-based modeling \citep{garrone2025adaptive}, and more directly on finite controller revision and machine-coached policy repair \citep{garrone2026coaching,garrone2026repairability}.

The primary research question is:

\begin{quote}
	\emph{When an adaptive agentic controller is represented through finite rules, explicit diagnostic predicates, explanation logs, and held-out re-evaluation, which classes of controller failure can be detected, locally repaired, or rejected without relying on unrestricted human-in-the-loop judgment?}
\end{quote}

The paper does not claim that all agentic failures can be detected or repaired. It asks whether specific controller-level failures can be made observable, classified, revised, and re-tested under a finite protocol. This position treats verification as an architectural property of the controller, not as an external audit applied only after deployment.

The proposed method is based on four elements. First, the controller is represented as a finite symbolic rule base with explicit priorities and conflicts. Second, numerical simulation states are converted into diagnostic predicates. Third, each controller action generates an explanation log recording observed predicates, applicable rules, selected rules, blocked rules, conflicts, priorities, and the chosen action. Fourth, when a failure is diagnosed, a finite edit library is used to revise the controller, and the revised controller is evaluated on held-out simulation runs.

The contribution is methodological. The paper provides a simulation-based verification protocol for finite adaptive controllers and reports benchmark experiments that separate detectable failures, locally repairable failures, and rejected repairs. It does not propose a new optimal controller, a general safety guarantee, or a replacement for human governance. Its purpose is to clarify which controller failures can be addressed through bounded, auditable, and reproducible revision operations.

Finally, this work is related to earlier research on adaptive, explainable, and contestable agent-based modeling \citep{garrone2025adaptive}, machine-coached policy revision \citep{garrone2026coaching}, and one-step repairability of finite symbolic controllers \citep{garrone2026repairability}. Its distinct contribution is the application of controller-level verification to the industrial agentic-AI deployment gap, together with an inventory-control benchmark that reports negative cases, rejected repairs, baseline comparisons, and one local smoothing-rule repair.

\section{Related Work}

The proposed framework connects several technical literatures that are usually treated separately: verification of finite-state and agentic systems, rule-based decision systems, defeasible reasoning, controller synthesis and repair, simulation-based verification, and inventory-control modeling. Its contribution is not to replace these traditions, but to combine a restricted subset of them into a controller-level protocol for adaptive agentic systems whose decisions can be represented, logged, revised, and re-tested.

\subsection{Verification, Monitoring, and Simulation Evidence}

Formal verification provides the strongest reference point for claims about system correctness. Model checking studies whether an abstract finite-state model satisfies a formal specification, typically expressed in temporal logic \citep{clarke1999}. This tradition is relevant because the present framework also treats verification as a relation between a system model and declared properties. However, the proposed protocol is weaker than full model checking. It does not exhaustively verify all reachable states. It evaluates controller behavior under declared simulation seeds, cloned initial states, or replicated stochastic conditions.

Runtime verification and monitoring provide a closer analogue. Runtime verification evaluates whether observed traces satisfy declared properties, rather than proving correctness over the full state space \citep{bauer2011}. The explanation log used here has a similar function: it records predicates, applicable rules, selected rules, blocked rules, priorities, and actions so that controller behavior can be inspected after execution. The difference is that the present framework links monitoring to finite rule revision. A trace is not only checked; it may also identify an admissible edit to the controller rule base.

Statistical model checking is also relevant because the proposed protocol relies on repeated stochastic simulation rather than exhaustive state exploration. Younes and Simmons \citep{younes2006} formulate probabilistic verification through statistical hypothesis testing over simulated executions. The present framework adopts a related evidential status: acceptance criteria are empirical and bounded. A repaired controller is accepted only relative to the declared seeds, stochastic assumptions, metrics, thresholds, and guardrails. This supports simulation evidence, not unrestricted deployment validity.

\subsection{Rule-Based Systems and Defeasible Reasoning}

The controller representation builds on the production-system tradition in artificial intelligence. Production systems represent decision logic through condition-action rules, and efficient pattern matching has long been a central implementation problem. The RETE algorithm is a canonical reference for matching many rules against many facts in production-system interpreters \citep{forgy1982}. The present framework does not depend on RETE specifically, but it shares the same basic assumption that explicit rule bodies can be matched against a current fact or predicate set.

The use of conflicts and priorities connects the framework to defeasible reasoning. Defeasible logic represents conclusions that may be overridden by stronger rules or priority relations. Antoniou et al. \citep{antoniou2001} provide a formal treatment of defeasible logic as a rule-based nonmonotonic reasoning system with priorities and transformations. This literature is relevant because the proposed controller does not treat all applicable rules as jointly executable. Instead, applicable rules may conflict, and a deterministic resolution operator selects the rule or rule set that controls the action.

The present framework remains narrower than general defeasible logic. It does not introduce a new nonmonotonic semantics. It uses a finite rule base, a finite predicate vocabulary, an explicit conflict relation, and a priority ordering as engineering objects for controller auditability and revision. The purpose is operational: to make controller decisions explainable, contestable, and locally revisable under simulation.

\subsection{Controller Synthesis, Repair, and Bounded Revision}

Controller synthesis studies how to construct a controller that enforces a desired behavioral specification. In discrete-event systems, Ramadge and Wonham \citep{ramadge1987} define supervisory control in terms of a plant, a language of admissible behavior, and a supervisor that restricts behavior to satisfy a specification. Reactive synthesis extends this idea to temporal-logic specifications and game-based construction of winning strategies; GR(1) synthesis is a tractable and widely used fragment for reactive designs \citep{bloem2012}.

The present paper is related to controller synthesis but does not solve a full synthesis problem. It assumes an existing finite controller and asks whether diagnosed failures can be repaired through a small edit library. The edit operations are deliberately restricted: rule addition, rule deletion, and priority revision. This makes the protocol closer to bounded controller repair than to unrestricted synthesis. The repaired controller is accepted only if the target failure improves and guardrail metrics remain within declared thresholds on held-out simulation runs.

This distinction is important for the scope of the contribution. A synthesis result would claim that a controller satisfying a specification can be constructed under formal assumptions. The present framework makes a weaker claim: when a failure is expressible in the predicate vocabulary, visible in explanation logs, and correctable by one admissible edit, the controller may be locally repairable under the declared simulation protocol.

The proposed protocol also sits near three more recent lines of work: runtime verification for self-adaptive systems, self-healing machine learning, and runtime governance of agentic actions. These literatures are important because they limit the novelty that can reasonably be claimed for monitoring, adaptation, or verification in isolation.

Runtime verification has already been integrated into self-adaptive feedback loops. For example, Carwehl et al. study runtime verification for self-adaptive systems whose requirements themselves may change, using dynamically adaptable monitors while preserving previously accumulated verification evidence \citep{carwehl2023}. The present paper therefore does not claim novelty from adding monitoring to an adaptive loop. Its narrower concern is whether a diagnosed controller-level failure can be mapped to a declared finite edit and whether the resulting controller change can be separately re-evaluated before acceptance.

Self-healing machine learning provides an even closer comparison. Rauba et al. propose a framework in which degradation is diagnosed and diagnosis-specific corrective actions are proposed and evaluated, including an LLM-based agentic implementation \citep{rauba2024}. The distinction sought here is consequently not ``diagnose and adapt'' by itself. The present protocol restricts the revisable object to an explicit finite controller, restricts repairs to a declared edit neighborhood, records the rule-level explanation trace that motivated the repair, and treats rejection or non-repairability as valid outcomes rather than assuming that adaptation should occur whenever degradation is detected.

Recent work on agent governance addresses a different but adjacent boundary. Palumbo et al. enforce declarative policies over real-world agentic systems through an external reference-monitor architecture \citep{palumbo2026}, while CAVA focuses on canonical action identity, approval binding, and attestable runtime action records across heterogeneous agent runtimes \citep{wang2026cava}. These approaches primarily ask whether a runtime action is permitted, attributable, or bound to an approval decision. The present paper instead studies a controller-change question: whether the decision policy itself should be modified after a diagnosed failure. This distinction can be summarized as
\[
\underbrace{V_a(a_t)}_{\text{action-level governance}}
\neq
\underbrace{V_R(R\rightarrow R')}_{\text{controller-change qualification}}.
\]
The two layers are complementary. Action-level governance can constrain what an agent may execute, while controller-change qualification can constrain how the logic that selects future actions may itself be revised.

\subsection{Inventory-Control Benchmarks}

The illustrative benchmark is positioned within inventory-control modeling, but it is not intended as a new inventory-optimization method. Classical inventory theory studies ordering policies under uncertain demand, cost, and replenishment constraints. Scarf's work on $(s,S)$ policies provides a canonical reference for dynamic inventory control under uncertainty \citep{scarf1960}. Clark and Scarf \citep{clark1960} extend inventory analysis to multi-echelon settings, where replenishment decisions propagate through linked inventory stages. Standard treatments of inventory management further distinguish stockout risk, holding cost, service level, replenishment timing, and capacity constraints \citep{silver1998,zipkin2000}.

The benchmark in this paper uses inventory control differently. It does not attempt to prove optimality of an ordering policy. Instead, inventory dynamics provide a transparent testbed for controller-level verification. Stockout, overstock, service-level loss, working-capital excess, order volatility, budget violation, and capacity violation are used as diagnostic properties. The purpose is to test whether a finite symbolic controller can make these failures observable, attribute them to rule-level decisions, revise the controller through finite edits, and re-evaluate the repair on held-out stochastic runs.

\subsection{Causal and Transfer-Learning Boundaries}

The framework also uses a limited causal vocabulary. Pearl's intervention framework is relevant because controller revision can be interpreted as an intervention on the decision rule rather than as a passive observation of system trajectories \citep{pearl2009}. However, the paper does not estimate causal effects from observational data. The causal claim is structural and methodological: rule edits define controlled changes to the controller mechanism, and before-after evaluation compares the resulting behavior under declared simulation conditions.

Similarly, the framework should not be confused with transfer learning. Transfer learning studies how knowledge from one domain, task, or distribution can improve performance in another \citep{pan2010}. The present paper does not claim that simulation evidence transfers automatically to deployment. Instead, it treats simulation-to-deployment shift as a failure condition. A controller that passes the simulation protocol may still fail under deployment-domain conditions if the simulation distribution differs from the operational distribution. This is why the framework supports pre-deployment qualification, not production-domain validity.

\section{Methodological Positioning and Scope}

This paper is a computational and methodological contribution. It produces a verification protocol, not empirical evidence about industrial deployment. The experimental environment is used as a controlled simulation benchmark. It is not used as a validated representation of a real industrial sector. This use of simulation follows the broader methodological tradition in agent-based modeling, where computational models are used to generate, inspect, and test mechanisms under declared assumptions rather than to provide direct empirical proof of deployment validity \citep{epstein1999,gilbert2008}. The reporting and reproducibility orientation is also consistent with the ODD tradition for describing agent-based and simulation models \citep{grimm2020}.

The paper produces four types of knowledge:

\begin{enumerate}[label=(\roman*)]
	\item a formal representation of adaptive agentic controllers as finite revisable rule systems;
	\item a diagnostic classification of controller-level failure modes;
	\item a finite repair protocol based on predefined rule-level edits;
	\item an evaluation design based on held-out simulation seeds or cloned initial states.
\end{enumerate}

The paper does not address cybersecurity, enterprise integration, legal liability, data-governance workflows, procurement, latency, or deployment infrastructure. It also does not address unrestricted agent autonomy. The scope is limited to controllers whose decision logic can be represented, logged, revised, and re-tested. Section~\ref{sec:industrial_instantiations} provides transfer scenarios
only; these examples add industrial interpretation but are not empirical
evidence that the protocol has been validated in those sectors.

\paragraph{Verification scope.}
The framework verifies only controller-level properties that satisfy four conditions: they are expressible in the declared predicate vocabulary, measurable through declared diagnostic metrics, attributable to logged controller decisions, and testable under the selected simulation protocol. It does not verify semantic truth of external information, cybersecurity, legal compliance, human organizational adoption, enterprise integration, or production-domain validity. These exclusions are not implementation details; they define the validity boundary of the contribution.
	
	\section{Enterprise Deployment: From Barriers to Verification Requirements}
	
	The practical motivation for the framework is the gap between agentic AI prototypes and production deployment. In consulting and enterprise settings, the relevant question is not whether an agent can perform a task once under demonstration conditions. The relevant question is whether its behavior can be observed, constrained, audited, and revised under operational uncertainty.
	
	\subsection{Cross-Platform Automation Pattern}
	
	A practical agentic-AI architecture should be reusable across client contexts while remaining compatible with the specific technology stack used in each organization. For this reason, the relevant design pattern is not tied to a single platform. It can be implemented with cloud services, enterprise workflow systems, RPA tools, CRM and ERP APIs, document-management systems, or internal applications. The common structure is that the agent interprets the task, retrieves trusted context, selects an allowed tool or workflow, and routes execution through governed systems.
	
	The central architectural principle is that the agent should not become the system of record. It should reason, coordinate, retrieve, and recommend, but authoritative execution should remain inside enterprise systems, APIs, workflow engines, RPA layers, or approval gates. This separation is important because it limits the operational authority of the agent and preserves existing governance boundaries.
	
	However, this pattern does not by itself solve the verification problem addressed in this paper. A governed execution layer can restrict what an agent is allowed to do, but it does not establish whether the controller's decision logic is observable, diagnostically adequate, stable under perturbation, or repairable after failure. The proposed framework addresses this residual problem by specifying a controller-level verification layer based on finite rules, diagnostic predicates, explanation logs, held-out evaluation, and bounded repair.
	
	\begin{figure}[h]
		\centering
		\resizebox{\textwidth}{!}{
			\begin{tikzpicture}[
				node distance=1.05cm,
				every node/.style={font=\small},
				stepbox/.style={
					rectangle,
					rounded corners,
					draw=black,
					thick,
					align=center,
					minimum width=5.2cm,
					minimum height=0.9cm,
					fill=gray!8
				},
				govbox/.style={
					rectangle,
					rounded corners,
					draw=black,
					thick,
					align=center,
					minimum width=5.2cm,
					minimum height=0.9cm,
					fill=blue!8
				},
				articlebox/.style={
					rectangle,
					rounded corners,
					draw=black,
					ultra thick,
					align=center,
					minimum width=6.0cm,
					minimum height=1.2cm,
					fill=yellow!25
				},
				riskbox/.style={
					rectangle,
					rounded corners,
					draw=red!70!black,
					thick,
					align=center,
					minimum width=6.0cm,
					minimum height=1.0cm,
					fill=red!8
				},
				arrow/.style={->, thick},
				dashedarrow/.style={->, thick, dashed}
				]
				
				\node[stepbox] (event) {
					\textbf{1. Business Event or User Request}\\
					Client process trigger, user query, workflow event
				};
				
				\node[stepbox, below=of event] (interpret) {
					\textbf{2. Agent Interprets the Task}\\
					Intent classification, task decomposition, routing
				};
				
				\node[stepbox, below=of interpret] (context) {
					\textbf{3. Agent Retrieves Trusted Context}\\
					Approved knowledge base, RAG, semantic layer, policies
				};
				
				\node[stepbox, below=of context] (toolselect) {
					\textbf{4. Agent Selects Allowed Tool or Workflow}\\
					Only tools exposed through governed permissions
				};
				
				\node[govbox, below=of toolselect] (execution) {
					\textbf{5. Execution Through Governed Layer}\\
					API, workflow engine, RPA bot, ERP, CRM, ticketing system
				};
				
				\node[govbox, below=of execution] (approval) {
					\textbf{6. Human Approval for High-Risk Cases}\\
					Escalation, review, sign-off, exception handling
				};
				
				\node[govbox, below=of approval] (logging) {
					\textbf{7. Logging, Evaluation, and Improvement}\\
					Logs, metrics, audits, tests, monitoring, feedback
				};
				
				\node[articlebox, right=2.0cm of approval] (article) {
					\textbf{Controller-level verification}\\
					finite rules, diagnostic predicates,\\
					explanation logs, held-out evaluation,\\
					bounded repair
				};
				
				\node[riskbox, right=2.0cm of toolselect] (gap) {
					\textbf{Deployment Verification Gap}\\
					The agent may appear capable,\\
					but its behavior may not be\\
					observable, bounded, or re-testable
				};
				
				\node[govbox, left=2.0cm of execution] (systemrecord) {
					\textbf{System-of-Record Principle}\\
					The agent does not become\\
					the authoritative database or\\
					execution authority
				};
				
				\draw[arrow] (event) -- (interpret);
				\draw[arrow] (interpret) -- (context);
				\draw[arrow] (context) -- (toolselect);
				\draw[arrow] (toolselect) -- (execution);
				\draw[arrow] (execution) -- (approval);
				\draw[arrow] (approval) -- (logging);
				
				\draw[dashedarrow] (toolselect.east) -- (gap.west);
				\draw[dashedarrow] (gap.south) -- (article.north);
				\draw[dashedarrow] (article.west) -- (approval.east);
				\draw[dashedarrow] (article.south west) -- (logging.east);
				
				\draw[dashedarrow] (systemrecord.east) -- (execution.west);
				\draw[dashedarrow] (systemrecord.north east) -- (toolselect.west);
				\draw[dashedarrow] (systemrecord.south east) -- (approval.west);
				
			\end{tikzpicture}
		}
		\caption{Cross-platform core design pattern for agent-enabled automation. The agent interprets, retrieves, coordinates, and recommends, while execution remains inside governed enterprise systems. The paper addresses the residual deployment verification gap by adding controller-level observability, diagnostics, held-out evaluation, and bounded repair.}
		\label{fig:cross_platform_core_pattern}
	\end{figure}
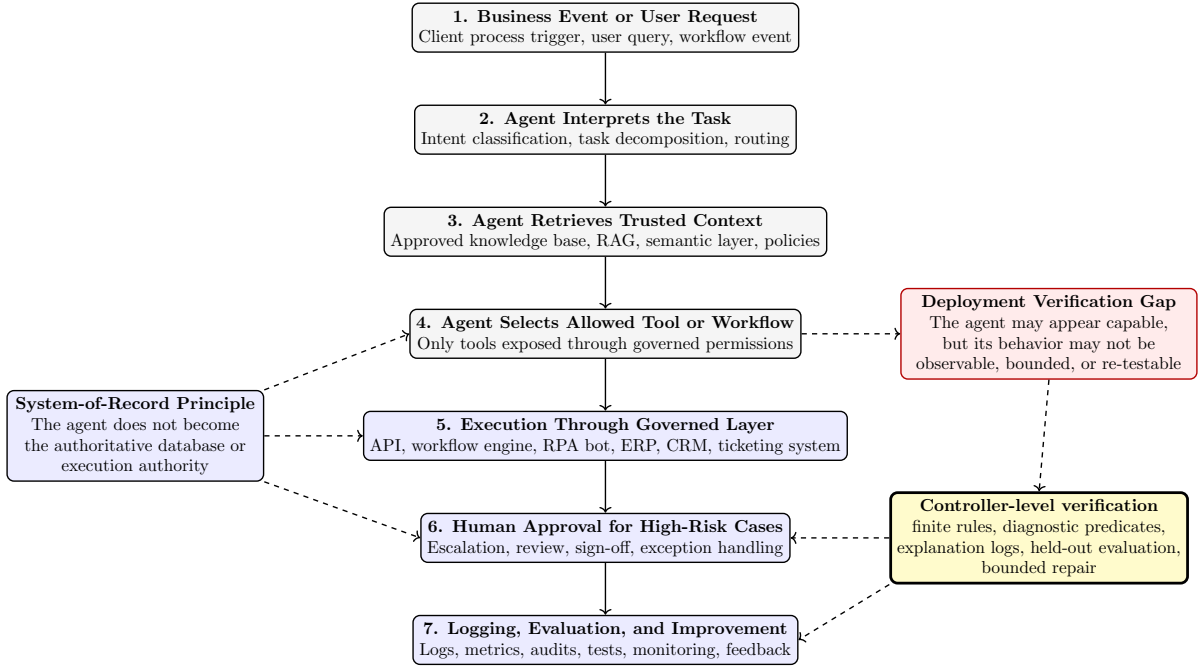
	
	This paper therefore translates common deployment barriers into controller-level verification requirements. The translation is not intended as a complete enterprise-AI governance framework. It is a technical narrowing of the deployment problem to the verification of adaptive controller behavior.
	
	\begin{table}[h]
		\centering
		\small
		\begin{tabular}{p{0.25\linewidth}p{0.34\linewidth}p{0.31\linewidth}}
			\toprule
			Deployment barrier & Verification requirement & Proposed mechanism \\
			\midrule
			Unobservable agent behavior 
			& The system must expose why an action was selected 
			& Explanation logs recording predicates, applicable rules, blocked rules, priorities, and actions \\
			
			Non-deterministic outputs 
			& The controller must be evaluated across repeated seeds or stochastic replications 
			& Held-out simulation seeds, cloned initial states, empirical quantile criteria \\
			
			Unsafe or unstable actions 
			& The controller must be checked against explicit guardrails 
			& Stockout rate, overstock rate, service-level loss, working-capital excess, and order-volatility thresholds \\
			
			Unrestricted human approval bottleneck 
			& Human judgment should be constrained to predefined review and revision operations where possible 
			& Finite edit library with rule addition, rule deletion, and priority revision \\
			
			Confidential or proprietary data 
			& Testing should avoid unnecessary exposure of operational data 
			& Synthetic or simulated environments, predicate abstraction, and aggregate diagnostic outputs \\
			
			Context-window and memory limits 
			& Operational decisions should not depend only on long implicit conversational context 
			& Finite predicate vocabulary, finite rule base, and bounded explanation objects \\
			
			Weak production qualification 
			& Prototype behavior must be compared against explicit acceptance criteria before deployment 
			& Before/after held-out evaluation and guardrail-preserving repair acceptance \\
			\bottomrule
		\end{tabular}
		\caption{Translation of agentic AI deployment barriers into controller-level verification requirements.}
		\label{tab:deployment_barriers}
	\end{table}
	
	For practitioners, the value of the framework is therefore not that it replaces enterprise governance, model-risk management, cybersecurity review, legal review, or system integration. Its value is narrower: it provides a technical testbed for evaluating whether an adaptive controller is observable, diagnosable, boundedly revisable, and empirically re-testable before operational use.
	
	\FloatBarrier
	\section{Industrial Instantiations of the Controller-Change Boundary}
	\label{sec:industrial_instantiations}
	
	The inventory benchmark is deliberately stylized, but the architectural problem is not inventory-specific. In many enterprise systems, several predictive or diagnostic sources feed a production decision policy, and the economically important question is not only whether the predictions are accurate. A second question is how the policy that uses those signals may be changed when evidence indicates degradation.
	
	The generic pattern is
	\[
	\text{multiple predictive or diagnostic signals}
	\rightarrow
	\boxed{\text{production decision policy }P_t}
	\rightarrow
	\text{business action}.
	\]
	When a controller-level diagnostic fails, the proposed framework inserts a bounded change-control step:
	\[
	D(P_t)=1
	\rightarrow
	P_t'=e(P_t),\quad e\in\mathcal{E}
	\rightarrow
	\text{independent re-evaluation}
	\rightarrow
	\{\text{retain},\text{accept},\text{reject},\text{escalate}\}.
	\]
	This section gives eight industrial instantiations. They are \emph{transfer scenarios}, not additional empirical validations of the present paper. The cited literatures already contain forecasting, optimization, ensemble, or adaptive methods for these domains. The intended contribution is therefore not to claim novelty in demand forecasting, credit scoring, ETA prediction, energy forecasting, or predictive maintenance. The purpose is to show where controller-change qualification could sit above those domain models and below authoritative enterprise execution.
	
	\begin{table}[t]
		\centering
		\scriptsize
		\begin{tabular}{p{0.14\linewidth}p{0.20\linewidth}p{0.18\linewidth}p{0.20\linewidth}p{0.17\linewidth}}
			\toprule
			Domain & Multiple signals & Controlled policy & Example bounded revision & Operational value / guardrail \\
			\midrule
			Demand planning & Statistical, sales, promotion, weather and foundation-model forecasts & Demand forecast used by planning & Reweight, recalibrate, exclude failing source & Inventory, service level, forecast stability \\
			Procurement & Demand, lead-time, cost and supplier-risk estimates & Order quantity / purchase timing & Add conservation rule, change source priority, retain incumbent & Working capital, shortage, supplier concentration \\
			Logistics / ETA & Neural, tree, traffic and carrier ETA estimates & ETA used by dispatch and promise logic & Exclude degraded source, switch combination rule & Fleet utilization, lateness, promise stability \\
			Energy & Weather and generation forecasts & Renewable forecast used in dispatch / balancing & Change model selection or fallback policy & Imbalance cost, reserve, forecast volatility \\
			Treasury & Macro, market, economist and scenario estimates & Hedge / budget recommendation policy & Change source weighting or conservative fallback & Hedge limits, liquidity, counterparty constraints \\
			Maintenance & Sensors, OEM models, anomaly detectors, prognostics & Inspection / maintenance trigger policy & Lower threshold, activate secondary model, escalate & Downtime, false alarms, missed failures \\
			Credit / risk & PD scores, bureau signals, behavioral models & Approval / limit recommendation policy & Activate challenger, change threshold, rollback & Loss, false negatives, fairness / policy constraints \\
			Workforce & Demand, absence, workload and operational forecasts & Staffing requirement policy & Change buffer, forecast source, escalation threshold & Labor cost, understaffing, service level \\
			\bottomrule
		\end{tabular}
		\caption{Illustrative industrial instantiations of the same controller-change qualification pattern. These scenarios are not empirical results of this paper.}
		\label{tab:industrial_instantiations}
	\end{table}
	
	\subsection{Demand Planning}
	
	Demand-planning systems already use ensemble and multi-model forecasting. Chien et al. combine demand-pattern classification, stacking, monitoring, and adaptive retraining for automotive after-market spare parts, and report reductions in forecast error and total cost in an industrial case \citep{chien2023}. More recent work also combines heterogeneous foundation-model forecasts to improve robustness across hierarchical retail-demand settings \citep{yang2025}. These studies show that forecasting and adaptation themselves are not the contribution sought here.
	
	The controller-change problem arises one level above the predictive models. Suppose a planning policy combines a statistical forecast, a commercial forecast, and a promotion-sensitive model. A persistent promotional bias may trigger several plausible responses: recalibrate one source, reduce its influence, exclude it temporarily, or retain the current policy because the evidence is insufficient. The finite-revision framework would require the diagnosis and permitted response set to be declared before a change is promoted. Acceptance would depend not only on forecast error but also on downstream guardrails such as service-level loss, inventory volatility, or working-capital exposure.
	
	\subsection{Procurement}
	
	AI and machine-learning methods are increasingly used across procurement and purchasing decision support, including demand planning, supplier selection, spend analysis, order management, and compliance \citep{balkan2025}. Forecasting and inventory control are also increasingly integrated rather than optimized as separate tasks; van der Haar et al., for example, use supervised learning to link feature-driven demand information directly to inventory decisions \citep{vanderhaar2024}.
	
	For controller-change qualification, the relevant object is the purchase policy rather than the demand forecast alone. Multiple signals---expected demand, supplier lead time, unit cost, liquidity, and supplier reliability---may feed a rule that determines order quantity and timing. If shortages rise, an unconstrained adaptive system may simply order more. A bounded controller instead distinguishes whether the failure is caused by an inaccurate source, an overly conservative rule, or a binding resource constraint. A proposed repair can then be rejected if it improves availability while violating working-capital, capacity, or supplier-concentration guardrails. This use case is particularly close to the financially constrained inventory benchmark used in the experiments.
	
	\subsection{Logistics and ETA}
	
	ETA prediction already benefits from heterogeneous model fusion. Huang et al. combine deep-network and tree-model predictions and show that complementary model classes can improve ETA accuracy and robustness \citep{huang2024}. This makes logistics a useful example precisely because the predictive fusion problem is already established.
	
	The additional controller-level question concerns what happens after one source or regime degrades. A dispatch policy may rely on ETA estimates from an internal model, a carrier estimate, and traffic-derived predictions. A detected deterioration can motivate several bounded edits: temporarily exclude one source, activate an alternative aggregation rule, or fall back to a conservative promise-time policy. The verifier can then compare the incumbent and challenger policies on later trips or historical replay while checking lateness, fleet-utilization, and promise-stability constraints. The agentic role is therefore policy maintenance, not ETA generation itself.
	
	\subsection{Energy Forecasting and Dispatch}
	
	Renewable-energy forecasting provides a strong example of why dynamic model selection is already prior art. Wu et al. use reinforcement learning to select forecasting models dynamically under changing wind and radiation conditions \citep{wu2025}; related work also constructs dynamic numerical-weather-prediction ensembles to avoid poorly performing members as conditions change \citep{dynamicnwp2024}. A claim of novelty based only on adaptive forecast selection would therefore be weak.
	
	The controller-change framework instead focuses on qualification of the adaptation. A production policy may use weather and generation forecasts to support balancing, reserve, or dispatch decisions. When a model's relative performance changes, a candidate policy edit can be proposed, but deployment would require evidence that the change improves the target metric without increasing forecast volatility, reserve stress, or other declared operational risks beyond threshold. This separates the ability to adapt from the authority to modify a production decision policy.
	
	\subsection{Treasury and Financial Planning}
	
	Financial forecasting likewise has a mature ensemble literature. For example, Wang et al. construct an exchange-rate ensemble specifically to exploit complementary predictive information from heterogeneous feature representations \citep{wang2021fx}. The present framework should therefore not be interpreted as a new financial-forecasting method.
	
	In an enterprise treasury setting, multiple macroeconomic, market, economist, and scenario forecasts may feed a hedge or budgeting recommendation. A candidate revision that improves point-forecast accuracy may still be inadmissible if it causes an excessive hedge ratio, liquidity use, concentration in a counterparty, or instability in budget assumptions. The controller-change layer makes this distinction explicit: predictive improvement is a diagnostic input, whereas policy admissibility is a multi-guardrail decision.
	
	\subsection{Predictive Maintenance}
	
	Predictive maintenance is naturally multi-model. Montero Jimenez et al. survey diagnostic and prognostic systems and emphasize that complex maintenance tasks often require complementary models rather than a single predictor \citep{montero2020}. Continual learning has also been proposed to adapt predictive-maintenance models under changing operating conditions \citep{hurtado2023}. These literatures again demonstrate that model combination and adaptation are not new by themselves.
	
	A controller-change formulation can instead focus on the intervention policy that consumes those models. Sensor models, OEM prognostics, and anomaly detectors may feed rules for inspection, maintenance, or escalation. When false negatives increase, possible edits include lowering an alarm threshold, activating a secondary diagnostic source, increasing inspection frequency, or escalating to human review. The repair should be accepted only if it improves the diagnosed failure while preserving constraints on false alarms, maintenance cost, downtime, and safety-relevant missed failures.
	
	\subsection{Credit and Risk Decision Support}
	
	Credit scoring already contains extensive ensemble and dynamic-selection research. Abdoli et al. benchmark dynamic ensemble-selection approaches on real credit-scoring data \citep{abdoli2020}, while broader surveys document both the predictive advantages and interpretability challenges of modern ensemble and machine-learning approaches in credit scoring \citep{moscato2021credit}.
	
	The controller-level application is therefore not to introduce another ensemble. Multiple probability-of-default estimates, bureau signals, behavioral models, and policy variables may feed an approval or credit-limit recommendation. A challenger policy can improve aggregate predictive accuracy while worsening false-negative risk or violating a declared segment-level policy constraint. Under the proposed protocol, this is a rejected repair rather than a successful optimization. Credit therefore illustrates why local predictive improvement and global controller acceptance must remain separate concepts.
	
	\subsection{Workforce Planning}
	
	Forecast-driven workforce planning provides a final example. Eichenseer et al. develop a machine-learning model for forecasting delivery positions in shop-floor logistics and use the resulting predictions as inputs to optimized workforce planning, reporting improvements over manual forecasting in the industrial case studied \citep{eichenseer2025}. This demonstrates the direct link from predictive accuracy to staffing decisions.
	
	A controller-change layer would govern the staffing policy rather than replace the forecasting model. Demand forecasts, absence estimates, workload signals, and contractual constraints may determine a staffing requirement or safety buffer. If the current policy begins to under-staff peak periods, a bounded repair may change a buffer, alter the preferred forecast source, or trigger an escalation rule. Held-out or later-period evaluation should then assess both labor cost and under-staffing/service-level risk. Authoritative schedule changes would remain within workforce-management systems and approval workflows.
	
	\subsection{Common Industrial Interpretation}
	
	Across these eight domains, the predictive layer and the controller-change layer should remain conceptually separate. Existing domain research already provides sophisticated forecasting, ensemble, optimization, and continual-learning methods. The proposed architecture does not compete with those methods. Instead, it treats them as signal generators or incumbent decision components whose use is governed by a finite, inspectable controller.
	
	The common industrial proposition is therefore:
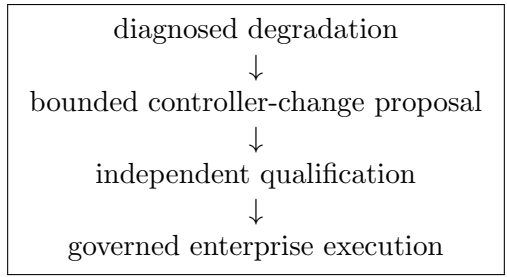
\[
\boxed{
	\begin{array}{c}
		\text{diagnosed degradation}
		\\
		\downarrow
		\\
		\text{bounded controller-change proposal}
		\\
		\downarrow
		\\
		\text{independent qualification}
		\\
		\downarrow
		\\
		\text{governed enterprise execution}
	\end{array}
}
\]
	The operational value of the framework lies at this policy-maintenance boundary. It is intended to make controller changes traceable and contestable while preserving existing systems of record, workflow controls, and human approval for high-risk cases. Empirical validation of these transfer scenarios remains future work; the experiments in this paper support only the inventory-control benchmark claims stated later in Section~\ref{sec:inventory_results}.
	
	\section{Problem Formulation}
	
	\subsection{Adaptive Agentic Controller}
	
	Let an adaptive agentic system operate in discrete time. At each time step $t$, the system has state
	\[
	s_t \in \mathcal{S},
	\]
	and the controller selects an action
	\[
	a_t \in \mathcal{A}.
	\]
	The environment then updates according to
	\[
	s_{t+1} = F(s_t,a_t,\zeta_t),
	\]
	where $\zeta_t$ denotes exogenous stochastic variation.
	
	The controller is called \emph{agentic} in the limited sense that it selects actions conditionally on observed state information, may update its intervention level over time, and can affect subsequent system states. The controller is called \emph{adaptive} when its actions or internal decision conditions depend on previous observations, diagnostic states, or feedback variables.
	
	\subsection{Finite Rule Representation}
	
	Let the controller be represented by a finite rule base
	\[
	R = \{r_1,\ldots,r_k\}.
	\]
	Each rule is a tuple
	\[
	r_j = (B_j,h_j,\pi_j),
	\]
	where $B_j$ is a finite set of body predicates, $h_j$ is a rule head or conclusion, and $\pi_j$ is a priority score.
	
	At time $t$, a predicate abstraction function
	\[
	A:\mathcal{S}\rightarrow 2^{\mathcal{P}}
	\]
	maps the numerical system state $s_t$ into a finite set of predicates
	\[
	P_t = A(s_t).
	\]
	A rule $r_j$ is applicable at time $t$ when
	\[
	B_j \subseteq P_t.
	\]
	The set of applicable rules is therefore
	\[
	R_t^{app} = \{r_j \in R : B_j \subseteq P_t\}.
	\]
	
	If rules conflict, the controller applies a deterministic resolution operator
	\[
	R_t^{sel} = \operatorname{Resolve}(R_t^{app},\perp,\pi),
	\]
	where $\perp$ denotes the conflict relation among rule heads and $\pi$ denotes the priority ordering. The selected rule set is then mapped to a controller action:
	\[
	a_t = H(R_t^{sel}).
	\]
	
	\subsection{Explanation Log}
	
	For each controller action, the system records an explanation object
	\[
	E_t =
	(P_t, R_t^{app}, R_t^{sel}, R_t^{blk}, \perp_t, \pi_t, a_t),
	\]
	where $R_t^{blk}$ denotes rules blocked by conflict or priority resolution. The explanation log over a simulation run is
	\[
	E_{0:T} = (E_0,\ldots,E_T).
	\]
	
	The explanation log is not treated as a natural-language justification. It is an operational trace of the controller's decision procedure. Its function is to identify which predicates, rules, conflicts, and priorities produced the controller action.

	\section{Verification Taxonomy}
	
	The paper distinguishes five verification layers. The distinction is necessary because different deployment failures require different forms of evidence. A controller may produce acceptable outputs while still having unstable behavior, inconsistent rule resolution, or poor robustness under stochastic variation.
	
	\subsection{Output Verification}
	
	Output verification checks whether observed outcomes satisfy declared numerical or categorical criteria. In the inventory-control benchmark, examples include stockout rate, overstock rate, service-level loss, working-capital excess, tracking error, and order volatility. Formally, output verification evaluates whether
	\[
	M_{out}(R,\omega) \leq \theta_{out},
	\]
	where $M_{out}$ is an output metric, $R$ is the controller rule base, $\omega$ is the simulation seed or stochastic condition, and $\theta_{out}$ is the acceptance threshold.
	
	Output verification is necessary but insufficient. It can detect that a controller failed, but it may not identify which rule, conflict, or priority relation generated the failure.
	
	\subsection{Behavioral Verification}
	
	Behavioral verification checks whether the controller follows acceptable trajectories, not only acceptable final or average outcomes. Let
	\[
	\tau_R(\omega) = (s_0,a_0,s_1,a_1,\ldots,s_T)
	\]
	be the trajectory induced by controller $R$ under condition $\omega$. Behavioral verification evaluates trajectory-level properties such as stockout episodes, overstock episodes, recurrence of working-capital excess, unstable replenishment, and order-allocation volatility.
	
	This layer is relevant when two controllers have similar average performance but differ in operational risk. For example, one controller may keep average inventory close to its target while generating frequent stockouts; another may avoid stockouts by holding excessive inventory and consuming unnecessary working capital.
	
	\subsection{Rule-Consistency Verification}
	
	Rule-consistency verification checks whether the symbolic controller remains executable and auditable after revision. A candidate rule base $R'$ passes this layer only if:
	
	\begin{enumerate}[label=(\roman*)]
		\item every rule has a unique identifier;
		\item every body predicate belongs to the declared predicate vocabulary;
		\item every rule head belongs to the declared action vocabulary;
		\item the conflict relation is symmetric;
		\item mutually exclusive actions are explicitly represented in the conflict relation;
		\item priorities are integer-valued or otherwise totally ordered;
		\item the resolution function returns a conflict-consistent selected set for all observed diagnostic states.
	\end{enumerate}
	
	Rule-consistency verification does not prove global correctness. It verifies that the revised controller remains syntactically valid, executable, and auditable.
	
	\subsection{Stochastic Robustness Verification}
	
	Stochastic robustness verification evaluates whether the controller satisfies acceptance criteria under repeated seeds or replications. Let $M(R,\omega_b)$ be a diagnostic metric under seed $\omega_b$, for $b=1,\ldots,B$. A conservative empirical criterion is:
	\[
	W_{\alpha,\theta}(R)=1
	\iff
	\hat{q}_{1-\alpha}
	\left(M(R,\omega_1),\ldots,M(R,\omega_B)\right)
	\leq \theta.
	\]
	
	This layer addresses non-determinism. A controller should not be classified as verified only because it passes under a favorable seed.
	
	\subsection{Confidentiality-Preserving Simulation}
	
	Confidentiality-preserving simulation evaluates controller behavior without requiring direct exposure of sensitive operational data. This can be done by replacing raw enterprise data with synthetic populations, simulated environments, aggregate constraints, or predicate-level abstractions. The relevant claim is limited: such simulation can support pre-production stress testing, but it does not eliminate the need for validation against deployment-domain evidence.
	
	\begin{table}[h]
		\centering
		\small
		\begin{tabular}{p{0.25\linewidth}p{0.34\linewidth}p{0.31\linewidth}}
			\toprule
			Verification layer & Main question & Evidence generated \\
			\midrule
			Output verification & Did the controller satisfy numerical thresholds? & Scalar diagnostic metrics \\
			Behavioral verification & Did the controller behave acceptably over time? & Trajectory motifs, episode lengths, transition patterns \\
			Rule-consistency verification & Is the revised controller executable and auditable? & Syntactic checks, conflict checks, priority checks \\
			Stochastic robustness verification & Does the result survive repeated seeds? & Replicated metrics, empirical quantiles, confidence intervals \\
			Confidentiality-preserving simulation & Can behavior be stress-tested without exposing raw data? & Synthetic or abstracted simulation traces \\
			\bottomrule
		\end{tabular}
		\caption{Verification taxonomy for adaptive agentic controllers.}
		\label{tab:verification_taxonomy}
	\end{table}
	
	\section{Controller-Level Failure Classes}
	
	Let $M(R,\omega)$ denote the vector of diagnostic metrics obtained when rule base $R$ is evaluated under simulation randomness or seed condition $\omega$. A controller failure is defined by a diagnostic predicate
	\[
	D_\ell(M(R,\omega)) = 1,
	\]
	where $D_\ell$ belongs to a finite failure vocabulary
	\[
	\mathcal{F} = \{f_1,\ldots,f_m\}.
	\]
	
	The paper focuses on controller-level failures rather than all possible system failures. The following classes are considered.
	
	\subsection{Stockout Failure}
	
	A stockout failure occurs when the controller repeatedly allows inventory to fall below the declared lower threshold. For inventory unit $i$, with inventory level $I_{i,t}$ and lower bound $L_i$, the stockout indicator is
	\[
	s_{i,t}^{-}=\mathbb{I}(I_{i,t}<L_i).
	\]
	The stockout rate is
	\[
	SR=(NT)^{-1}\sum_{i=1}^{N}\sum_{t=1}^{T}s_{i,t}^{-}.
	\]
	A stockout failure is diagnosed when
	\[
	SR>\theta_{SR}.
	\]
	
	\subsection{Overstock Failure}
	
	An overstock failure occurs when the controller repeatedly allows inventory to exceed the declared upper threshold. For upper bound $U_i$, the overstock indicator is
	\[
	s_{i,t}^{+}=\mathbb{I}(I_{i,t}>U_i).
	\]
	The overstock rate is
	\[
	OR=(NT)^{-1}\sum_{i=1}^{N}\sum_{t=1}^{T}s_{i,t}^{+}.
	\]
	An overstock failure is diagnosed when
	\[
	OR>\theta_{OR}.
	\]
	
	\subsection{Service-Level Failure}
	
	A service-level failure occurs when realized demand cannot be met by available inventory and replenishment. Let unmet demand be
	\[
	S_{i,t}=\max\{0,D_{i,t}-(I_{i,t}+Q_{i,t})\}.
	\]
	Service-level loss is
	\[
	SL=T^{-1}\sum_{t=1}^{T}\sum_{i=1}^{N}S_{i,t}.
	\]
	A service-level failure is diagnosed when
	\[
	SL>\theta_{SL}.
	\]
	
	\subsection{Working-Capital Failure}
	
	A controller may avoid shortages by holding excessive inventory. Let working-capital excess be
	\[
	WCE=T^{-1}\sum_{t=1}^{T}\sum_{i=1}^{N}c_i\max(0,I_{i,t}-U_i),
	\]
	where $c_i$ is the unit replenishment cost. A working-capital failure is diagnosed when
	\[
	WCE>\theta_{WCE}.
	\]
	
	\subsection{Order-Volatility Failure}
	
	Let $Q_{i,t}$ denote the replenishment allocation for unit $i$. Order volatility is defined as
	\[
	OV=(N(T-1))^{-1}\sum_{i=1}^{N}\sum_{t=2}^{T}|Q_{i,t}-Q_{i,t-1}|.
	\]
	A volatility failure is diagnosed when
	\[
	OV>\theta_{OV}.
	\]
	
	\subsection{Tracking Failure}
	
	If the controller is expected to maintain inventory near target level $I_i^\ast$, tracking error is defined as
	\[
	TE=(NT)^{-1}\sum_{i=1}^{N}\sum_{t=1}^{T}|I_{i,t}-I_i^\ast|.
	\]
	A tracking failure is diagnosed when
	\[
	TE>\theta_{TE}.
	\]
	
	\subsection{Budget-Constraint Failure}
	
	A budget-constraint failure occurs when the replenishment allocation exceeds the available financial envelope. If $B_t$ is the available budget or liquidity at time $t$, the budget violation rate is
	\[
	BVR=T^{-1}\sum_{t=1}^{T}\mathbb{I}\left(\sum_i c_iQ_{i,t}>B_t\right).
	\]
	A budget-constraint failure is diagnosed when
	\[
	BVR>\theta_{BVR}.
	\]
	
	\subsection{Capacity-Constraint Failure}
	
	A capacity-constraint failure occurs when the replenishment allocation exceeds available operational capacity. If $K_t$ is the available replenishment capacity at time $t$, the capacity violation rate is
	\[
	CVR=T^{-1}\sum_{t=1}^{T}\mathbb{I}\left(\sum_i Q_{i,t}>K_t\right).
	\]
	A capacity-constraint failure is diagnosed when
	\[
	CVR>\theta_{CVR}.
	\]
	
	\subsection{Conflict-Resolution Failure}
	
	A conflict-resolution failure occurs when the explanation log shows that a desirable corrective rule was applicable but repeatedly blocked by a higher-priority rule. Let $b_j$ denote the number of times rule $r_j$ is blocked while applicable. A conflict-resolution failure is diagnosed when
	\[
	b_j > \theta_B
	\]
	and the associated diagnostic metric remains outside the accepted range.
	
	\section{Finite Repair Protocol}
	
	\subsection{Edit Library}
	
	A repair operation is a finite edit applied to the controller rule base. Let
	\[
	\mathcal{E} = \mathcal{E}_{add} \cup \mathcal{E}_{del} \cup \mathcal{E}_{prio}
	\]
	be a predefined edit library.
	
	An addition edit adds a rule:
	\[
	e_{add}(R) = R \cup \{r^\ast\}.
	\]
	
	A deletion edit removes a rule:
	\[
	e_{del}(R) = R \setminus \{r_j\}.
	\]
	
	A priority edit changes a priority value:
	\[
	e_{prio}(R) = (R \setminus \{r_j\}) \cup \{(B_j,h_j,\pi_j')\}.
	\]
	
	The one-step edit neighborhood of rule base $R$ is
	\[
	\Gamma(R) = \{e(R): e\in\mathcal{E}, R\in\operatorname{dom}(e)\}.
	\]
	
	\subsection{Template-Constrained Coaching}
	
	A coaching function maps diagnostics and explanation logs to admissible edits:
	\[
	\Gamma_c(D,E_{0:T},R) \rightarrow e \in \mathcal{E} \cup \{\varnothing\}.
	\]
	The function is template-constrained. It does not accept arbitrary natural-language advice. It may only return predefined edits.
	
	For example, if service-level loss is diagnosed and the explanation log shows that stockout-risk predicates are repeatedly observed without prioritizing replenishment, the coaching function may add the following rule:
	\[
	\text{IF stockout\_risk AND budget\_available THEN prioritize\_stockout\_risk.}
	\]
	
	\subsection{Working Predicate}
	
	Let $W(R)$ denote whether the controller satisfies the accepted diagnostic and guardrail criteria. Under fixed seeds, $W(R)$ may be defined deterministically:
	\[
	W(R)=1 \iff M(R,\omega_0)\leq \theta.
	\]
	
	Under stochastic evaluation, a replicated version is preferable:
	\[
	W_{\alpha,\theta}(R)=1
	\iff
	\Pr_{\omega}\{M(R,\omega)\leq \theta\}\geq 1-\alpha.
	\]
	With a finite replication set $\omega_1,\ldots,\omega_B$, this may be implemented using an empirical quantile:
	\[
	W_{\alpha,\theta}(R)=1
	\iff
	\hat{q}_{1-\alpha}
	\left(
	M(R,\omega_1),\ldots,M(R,\omega_B)
	\right)
	\leq \theta.
	\]
	
	\subsection{One-Step Repairability}
	
	A controller failure is one-step repairable when the current controller fails but at least one admissible edit restores the working predicate:
	\[
	W(R)=0
	\quad\text{and}\quad
	\exists R'\in\Gamma(R): W(R')=1.
	\]
	
	This definition does not imply general repairability. It only states that one edit from the declared finite library is sufficient under the declared evaluation protocol.
	
	\section{Verification Protocol}
	
	The verification procedure is as follows.
	
\begin{enumerate}[label=\textbf{Step \arabic*.}, leftmargin=2.5cm]
	\item Declare the controller rule base $R$, predicate vocabulary $\mathcal{P}$, conflict relation $\perp$, and priority ordering $\pi$.
	\item Declare the diagnostic metrics, failure thresholds, and guardrail constraints.
	\item Run the controller under training or diagnostic simulation seeds.
	\item Generate explanation logs for all controller actions.
	\item Diagnose failure classes using the finite failure vocabulary $\mathcal{F}$.
	\item Apply the template-constrained coaching function to select one admissible edit.
	\item Construct the revised rule base $R'$.
	\item Evaluate $R$ and $R'$ on held-out seeds or cloned initial states.
	\item Accept the repair only if the targeted failure improves and guardrail constraints remain satisfied.
\end{enumerate}
	
	\noindent The protocol separates diagnosis from evaluation. A repair is proposed using diagnostic runs but assessed on held-out runs. This separation reduces the risk that the controller is merely tuned to the failure instance used to generate the edit.
	
	\section{Baselines}
	
	Repair results should be interpreted against explicit baselines.
	
	\subsection{No-Repair Baseline}
	
	The no-repair baseline evaluates the original failed controller without modification. It measures whether failure persists in held-out evaluation:
	\[
	R' = R.
	\]
	
	\subsection{Random-Edit Baseline}
	
	The random-edit baseline selects one admissible edit from $\Gamma(R)$ according to a declared randomization rule. This tests whether the coaching template performs better than arbitrary rule-base modification.
	
	\subsection{Original-Rule Restoration Baseline}
	
	If failures are generated by rule removal, an original-rule restoration baseline allows only one removed rule to be restored:
	\[
	\mathcal{E}_{restore} =
	\{e_j : R \mapsto R\cup\{r_j\}, r_j\in R_0\setminus R\}.
	\]
	
	\subsection{Numeric Controller Baseline}
	
	A numeric controller updates replenishment allocations directly through a fixed equation rather than explicit rules. This baseline tests whether symbolic contestability adds diagnostic value relative to a non-symbolic adaptive controller.
	
	\subsection{Manual-Tuning Proxy}
	
	A manual-tuning proxy applies a prespecified parameter adjustment without explanation-log-based rule revision. This baseline distinguishes auditable controller repair from ordinary parameter tuning.
	
\section{Illustrative Inventory-Control Benchmark}
\label{sec:inventory_benchmark}

The experimental component uses a stylized inventory-control simulation. The purpose is not to forecast inventory investment or optimize supply-chain policy. The purpose is to provide a controlled environment in which controller-level failures can be generated, diagnosed, repaired, and re-tested.

The benchmark is formulated as a financially constrained stock-adjustment problem with two-sided risk. There are $N$ inventory units, indexed by $i=1,\ldots,N$. These units may represent products, SKUs, stores, warehouses, business units, or demand nodes. At time $t$, unit $i$ has inventory level $I_{i,t}$, demand $D_{i,t}$, replenishment allocation $Q_{i,t}$, lower inventory threshold $L_i$, upper inventory threshold $U_i$, and target inventory level $I_i^\ast$.

Inventory evolves according to:

\[
I_{i,t+1}
=
\max\{0,I_{i,t}+Q_{i,t}-D_{i,t}\}.
\]

Demand is stochastic:

\[
D_{i,t}
=
\max\{0,\bar{D}_i+\epsilon_{i,t}\},
\]

where $\bar{D}_i$ is baseline demand and $\epsilon_{i,t}$ is an exogenous demand shock.

The controller must allocate replenishment subject to a capacity constraint,

\[
\sum_{i=1}^{N} Q_{i,t}\leq K_t,
\]

and a financial constraint,

\[
\sum_{i=1}^{N} c_i Q_{i,t}\leq B_t,
\]

where $K_t$ is available replenishment capacity, $B_t$ is available budget or liquidity, and $c_i$ is the unit replenishment cost.

The relevant operational risk is two-sided. A shortage failure occurs when inventory falls below the lower bound,

\[
I_{i,t}<L_i,
\]

whereas an overstock failure occurs when inventory exceeds the upper bound,

\[
I_{i,t}>U_i.
\]

The controller therefore must not simply increase or reduce replenishment. It must allocate limited capacity and budget across competing units while avoiding stockouts, overstock, excessive working-capital use, and unstable ordering behavior.

The symbolic controller observes predicates such as:

\[
I_{i,t}<L_i
\Rightarrow
\texttt{stockout\_risk}(i),
\]

\[
I_{i,t}>U_i
\Rightarrow
\texttt{overstock\_risk}(i),
\]

\[
\sum_i c_i Q_{i,t-1}\geq \kappa_B B_{t-1}
\Rightarrow
\texttt{budget\_binding},
\]

\[
\sum_i Q_{i,t-1}\geq \kappa_K K_{t-1}
\Rightarrow
\texttt{capacity\_binding},
\]

\[
|Q_{i,t-1}-Q_{i,t-2}|>\theta_Q
\Rightarrow
\texttt{order\_volatility\_high}(i).
\]

The controller selects actions such as:

\[
\begin{aligned}
\mathcal{A}=\{&
\texttt{increase\_order},
\texttt{reduce\_order},
\texttt{hold\_order},\\
&
\texttt{prioritize\_stockout\_risk},
\texttt{smooth\_order},
\texttt{delay\_low\_priority\_order}
\}.
\end{aligned}
\]

This environment is used as an agentic controller verification benchmark. It is not a calibrated inventory model and is not intended to predict firm-level inventory investment. Its function is to test whether specific controller-level failures can be detected, attributed to controller logic, repaired through finite edits, and re-evaluated under controlled stochastic variation.

The benchmark has four design requirements.

\begin{enumerate}[label=(\roman*)]
	\item \textbf{Finite controller representation.} The controller must be represented by a finite rule base, finite predicate vocabulary, finite action vocabulary, explicit conflict relation, and deterministic priority-resolution procedure.
	
	\item \textbf{Two-sided failure classes.} The benchmark must specify shortage and excess-inventory failures before evaluation, together with working-capital, capacity, volatility, tracking, and conflict-resolution failures.
	
	\item \textbf{Explicit acceptance criteria.} Each failure class must have a numerical threshold or logical condition that determines whether the controller is classified as working.
	
	\item \textbf{Held-out repair evaluation.} A repair may be proposed using diagnostic seeds, but it must be accepted only after comparison between the original and revised controller on held-out seeds or cloned initial states.
\end{enumerate}

The benchmark is therefore not designed to show that the controller is optimal. It is designed to test whether financially constrained stock-adjustment failures can be made observable, diagnosable, boundedly repairable, and empirically re-tested.

	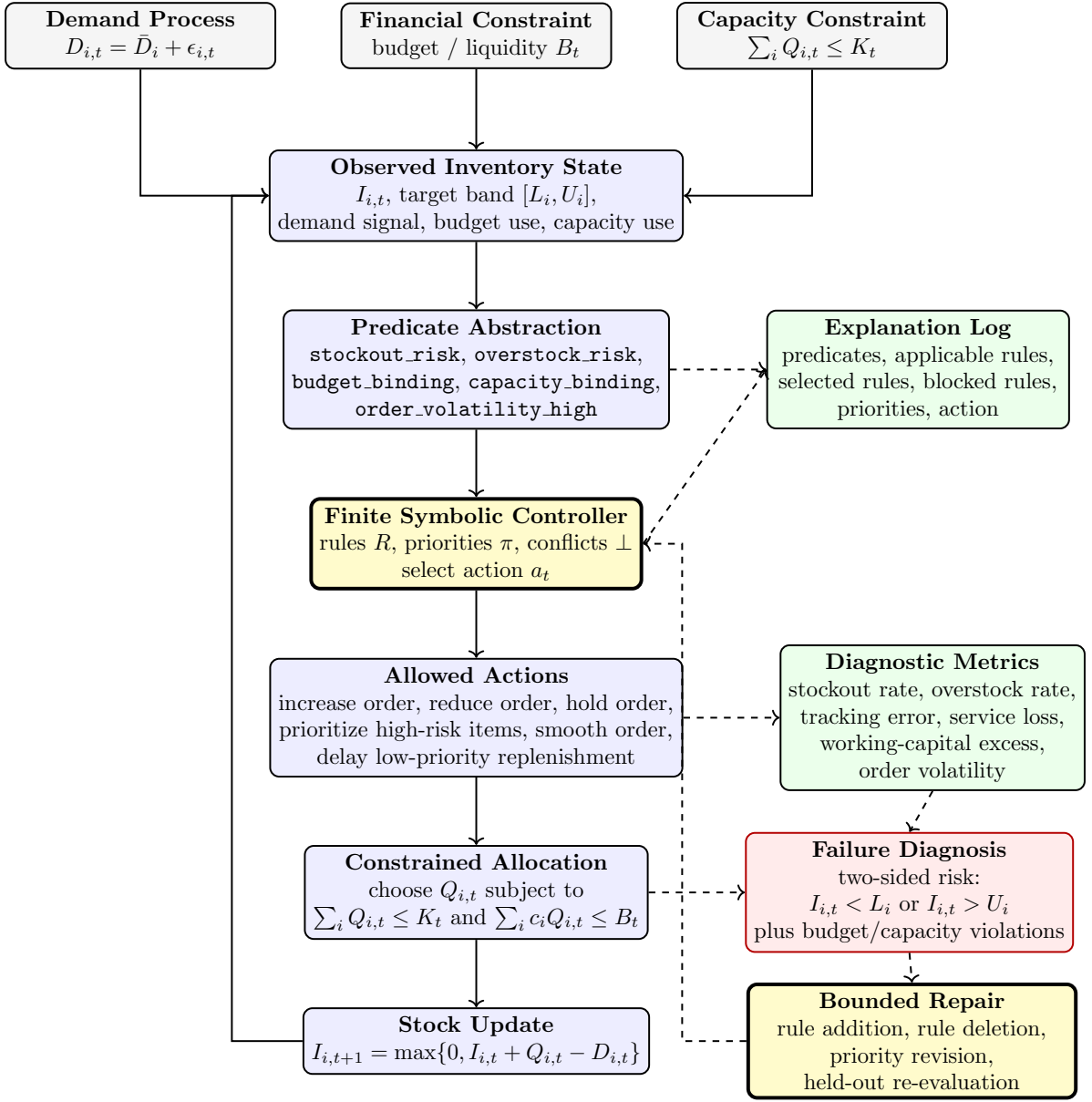
\begin{figure}[!htbp]
	\centering
	\resizebox{\textwidth}{!}{
		\begin{tikzpicture}[
			node distance=1.05cm,
			every node/.style={font=\small},
			statebox/.style={
				rectangle,
				rounded corners,
				draw=black,
				thick,
				align=center,
				minimum width=4.2cm,
				minimum height=0.9cm,
				fill=gray!8
			},
			processbox/.style={
				rectangle,
				rounded corners,
				draw=black,
				thick,
				align=center,
				minimum width=4.6cm,
				minimum height=0.95cm,
				fill=blue!7
			},
			rulebox/.style={
				rectangle,
				rounded corners,
				draw=black,
				ultra thick,
				align=center,
				minimum width=5.1cm,
				minimum height=1.1cm,
				fill=yellow!25
			},
			metricbox/.style={
				rectangle,
				rounded corners,
				draw=black,
				thick,
				align=center,
				minimum width=4.7cm,
				minimum height=0.95cm,
				fill=green!8
			},
			failbox/.style={
				rectangle,
				rounded corners,
				draw=red!70!black,
				thick,
				align=center,
				minimum width=4.7cm,
				minimum height=0.95cm,
				fill=red!8
			},
			arrow/.style={->, thick},
			dashedarrow/.style={->, thick, dashed}
			]
			
			\node[statebox] (demand) {
				\textbf{Demand Process}\\
				$D_{i,t}=\bar{D}_i+\epsilon_{i,t}$
			};
			
			\node[statebox, right=1.0cm of demand] (finance) {
				\textbf{Financial Constraint}\\
				budget / liquidity $B_t$
			};
			
			\node[statebox, right=1.0cm of finance] (capacity) {
				\textbf{Capacity Constraint}\\
				$\sum_i Q_{i,t}\leq K_t$
			};
			
			\node[processbox, below=1.25cm of finance] (state) {
				\textbf{Observed Inventory State}\\
				$I_{i,t}$, target band $[L_i,U_i]$,\\
				demand signal, budget use, capacity use
			};
			
			\node[processbox, below=of state] (predicates) {
				\textbf{Predicate Abstraction}\\
				\texttt{stockout\_risk}, \texttt{overstock\_risk},\\
				\texttt{budget\_binding}, \texttt{capacity\_binding},\\
				\texttt{order\_volatility\_high}
			};
			
			\node[rulebox, below=of predicates] (controller) {
				\textbf{Finite Symbolic Controller}\\
				rules $R$, priorities $\pi$, conflicts $\perp$\\
				select action $a_t$
			};
			
			\node[processbox, below=of controller] (actions) {
				\textbf{Allowed Actions}\\
				increase order, reduce order, hold order,\\
				prioritize high-risk items, smooth order,\\
				delay low-priority replenishment
			};
			
			\node[processbox, below=of actions] (allocation) {
				\textbf{Constrained Allocation}\\
				choose $Q_{i,t}$ subject to\\
				$\sum_i Q_{i,t}\leq K_t$ and $\sum_i c_iQ_{i,t}\leq B_t$
			};
			
			\node[processbox, below=of allocation] (update) {
				\textbf{Stock Update}\\
				$I_{i,t+1}=\max\{0,I_{i,t}+Q_{i,t}-D_{i,t}\}$
			};
			
			\node[metricbox, right=1.5cm of predicates] (logs) {
				\textbf{Explanation Log}\\
				predicates, applicable rules,\\
				selected rules, blocked rules,\\
				priorities, action
			};
			
			\node[metricbox, right=1.5cm of actions] (metrics) {
				\textbf{Diagnostic Metrics}\\
				stockout rate, overstock rate,\\
				tracking error, service loss,\\
				working-capital excess,\\
				order volatility
			};
			
			\node[failbox, right=1.5cm of allocation] (failure) {
				\textbf{Failure Diagnosis}\\
				two-sided risk:\\
				$I_{i,t}<L_i$ or $I_{i,t}>U_i$\\
				plus budget/capacity violations
			};
			
			\node[rulebox, right=1.5cm of update] (repair) {
				\textbf{Bounded Repair}\\
				rule addition, rule deletion,\\
				priority revision,\\
				held-out re-evaluation
			};
			
			\draw[arrow] (demand.south) |- (state.west);
			\draw[arrow] (finance.south) -- (state.north);
			\draw[arrow] (capacity.south) |- (state.east);
			
			\draw[arrow] (state) -- (predicates);
			\draw[arrow] (predicates) -- (controller);
			\draw[arrow] (controller) -- (actions);
			\draw[arrow] (actions) -- (allocation);
			\draw[arrow] (allocation) -- (update);
			
			\draw[arrow] (update.west) -- ++(-1.1,0) |- (state.west);
			
			\draw[dashedarrow] (predicates.east) -- (logs.west);
			\draw[dashedarrow] (controller.east) -- (logs.west);
			\draw[dashedarrow] (actions.east) -- (metrics.west);
			\draw[dashedarrow] (allocation.east) -- (failure.west);
			\draw[dashedarrow] (metrics.south) -- (failure.north);
			\draw[dashedarrow] (failure.south) -- (repair.north);
			\draw[dashedarrow] (repair.west) -- ++(-1.0,0) |- (controller.east);
			
		\end{tikzpicture}
	}
	\caption{Inventory benchmark implemented as a financially constrained stock-adjustment and allocation problem. The controller observes inventory, demand, budget, and capacity states; maps numerical states into diagnostic predicates; selects admissible replenishment actions through finite rules; allocates limited replenishment under capacity and financial constraints; and updates inventory dynamically. The verification layer records explanation logs, computes two-sided diagnostic metrics, diagnoses failures, and evaluates bounded rule repairs on held-out simulation seeds.}
	\label{fig:inventory_benchmark}
\end{figure}
	
\section{Evaluation Metrics}
\label{sec:inventory_metrics}

The main evaluation metrics are reported before interpretation.

\begin{table}[h]
	\centering
	\small
	\renewcommand{\arraystretch}{1.25}
	\begin{tabular}{lll}
		\toprule
		Metric & Definition & Failure Class \\
		\midrule
		Stockout rate
		&
		$SR=(NT)^{-1}\sum_i\sum_t \mathbb{I}(I_{i,t}<L_i)$
		&
		Shortage failure \\
		
		Overstock rate
		&
		$OR=(NT)^{-1}\sum_i\sum_t \mathbb{I}(I_{i,t}>U_i)$
		&
		Excess-inventory failure \\
		
		Service-level loss
		&
		$SL=T^{-1}\sum_t\sum_i \max(0,D_{i,t}-(I_{i,t}+Q_{i,t}))$
		&
		Unmet-demand failure \\
		
		Working-capital excess
		&
		$WCE=T^{-1}\sum_t\sum_i c_i\max(0,I_{i,t}-U_i)$
		&
		Working-capital failure \\
		
		Tracking error
		&
		$TE=(NT)^{-1}\sum_i\sum_t |I_{i,t}-I_i^\ast|$
		&
		Target-tracking failure \\
		
		Order volatility
		&
		$OV=(N(T-1))^{-1}\sum_i\sum_{t=2}^T |Q_{i,t}-Q_{i,t-1}|$
		&
		Volatility failure \\
		
		Budget violation rate
		&
		$BVR=T^{-1}\sum_t \mathbb{I}(\sum_i c_iQ_{i,t}>B_t)$
		&
		Budget-constraint failure \\
		
		Capacity violation rate
		&
		$CVR=T^{-1}\sum_t \mathbb{I}(\sum_i Q_{i,t}>K_t)$
		&
		Capacity-constraint failure \\
		
		Blocked-rule count
		&
		$b_j$
		&
		Conflict-resolution failure \\
		
		\bottomrule
	\end{tabular}
	\caption{Inventory-control diagnostics and associated failure classes.}
	\label{tab:inventory_diagnostics}
\end{table}

A repair is locally successful for a metric $m$ only if the held-out evaluation statistic for that metric improves and falls below its declared threshold. A repair is globally accepted only if all target failures diagnosed for the repaired controller are restored below threshold, all guardrails remain within threshold, and the revised rule base remains syntactically valid. For example, a stockout repair is not accepted merely because $SR$ or $SL$ decreases. It is accepted only if the relevant target threshold is restored while $OR$, $WCE$, $OV$, $BVR$, $CVR$, and the conflict metric remain within accepted bounds. This distinction between local repair and global acceptance is used in the experiments below.

\section{Failure Classes and Repair Expectations}
\label{sec:inventory_expected_results}

Before reporting the experiments, Table~\ref{tab:inventory_failureclasses} states the diagnostic expectations used to structure the benchmark. The expectation is not that all inventory-control failures are repairable. The expectation is that failures can be classified by whether they are detectable, whether they are plausibly one-step repairable under the finite edit library, and whether they require predicate redesign, template expansion, or domain-level changes.

\begin{table}[h]
	\centering
	\small
		\begin{tabular}{p{0.25\linewidth}p{0.14\linewidth}p{0.22\linewidth}p{0.27\linewidth}}
		\toprule
		Failure class & Detectable? & Repairable? & Typical edit \\
		\midrule
		Stockout failure & Yes & Sometimes & Add or prioritize replenishment rule \\
		Overstock failure & Yes & Sometimes & Add or prioritize reduction rule \\
		Service-level loss & Yes & Sometimes & Prioritize high-risk units \\
		Working-capital excess & Yes & Sometimes & Add budget-conservation rule \\
		Order volatility & Yes & Sometimes & Add smoothing rule or priority edit \\
		Tracking failure & Yes & Sometimes & Add target-band rule \\
		Budget-constraint failure & Yes & Often, if allocation is projected & Revise feasibility rule \\
		Capacity-constraint failure & Yes & Often, if allocation is projected & Revise allocation rule \\
		Conflict-resolution failure & Yes, via logs & Often, if conflict is local & Priority edit \\
		Missing predicate failure & No, not directly & No, unless vocabulary expands & Predicate redesign required \\
		Distribution-shift failure & Partially & Not guaranteed & Re-evaluation or new templates \\
		\bottomrule
	\end{tabular}
	\caption{Diagnostic expectations for inventory-control failure classes before held-out evaluation.}
	\label{tab:inventory_failureclasses}
\end{table}

The two-sided nature of the benchmark is important. A repair that reduces stockouts may increase overstock or working-capital excess. Conversely, a repair that reduces overstock may increase service-level loss. The benchmark therefore requires guardrail-preserving repair rather than single-metric improvement.
	
	\section{Negative Results and Failure Conditions}
	
	A verification protocol is informative only if it identifies where it fails. The proposed architecture is expected to fail under at least five conditions.
	
	\subsection{Missing-Predicate Failure}
	
	A missing-predicate failure occurs when the controller's predicate vocabulary cannot express the relevant diagnostic condition. Let $\mathcal{P}$ be the declared predicate vocabulary and let $f^\ast$ be a failure condition. If no predicate composition over $\mathcal{P}$ can represent $f^\ast$, then the controller cannot diagnose the failure through the symbolic layer.
	
	In this case, rule revision is insufficient. Adding a rule over the existing vocabulary cannot repair a failure that the vocabulary cannot represent. The required intervention is predicate redesign:
	\[
	\mathcal{P} \rightarrow \mathcal{P}'.
	\]
	
	This is a structural limitation. The controller may be fully auditable with respect to its declared predicates while remaining blind to an undeclared failure mode.
	
	\subsection{Inadequate-Template Failure}
	
	An inadequate-template failure occurs when the failure is detectable, but the edit library lacks an appropriate correction. Formally, for a failing controller $R$:
	\[
	W(R)=0
	\quad\text{and}\quad
	\forall R'\in\Gamma(R),\; W(R')=0.
	\]
	The failure is visible but not one-step repairable. This case is important because it separates observability from repairability. A controller may explain why it failed without possessing a valid predefined edit for correction.
	
	\subsection{Rule-Conflict Failure}
	
	A rule-conflict failure occurs when a corrective rule is added but is blocked by existing higher-priority rules or by the conflict-resolution procedure. Let $r_c$ be a corrective rule and let $b_c$ be its blocked count while applicable. If
	\[
	b_c > \theta_B
	\]
	and the target failure persists, then the repair failed because the controller's priority structure prevented the corrective action from being selected.
	
	This failure can sometimes be repaired through priority revision. However, if the conflict relation itself is misspecified, changing priorities may only move the failure from one rule to another.
	
	\subsection{Stochastic-Instability Failure}
	
	A stochastic-instability failure occurs when the repair succeeds under diagnostic seeds but fails under held-out seeds. Let $R'$ be the revised controller. The repair is unstable when:
	\[
	W(R')=1 \quad \text{on diagnostic seeds}
	\]
	but
	\[
	W_{\alpha,\theta}(R')=0 \quad \text{on held-out seeds}.
	\]
	This failure indicates that the repair may have overfit a local stochastic realization rather than corrected a stable controller defect.
	
	\subsection{Simulation-to-Deployment Shift Failure}
	
	A simulation-to-deployment shift failure occurs when the controller satisfies the verification protocol in simulation but fails under deployment-domain conditions. Let $\mathcal{D}_{sim}$ denote the simulation distribution and $\mathcal{D}_{dep}$ the deployment distribution. If
	\[
	\mathcal{D}_{sim} \neq \mathcal{D}_{dep},
	\]
	then verification under $\mathcal{D}_{sim}$ does not imply verification under $\mathcal{D}_{dep}$.
	
	The framework can reduce this risk through sensitivity analysis, scenario variation, synthetic populations, and held-out testing. It cannot eliminate the risk without deployment-domain validation.
	
	\begin{table}[h]
		\centering
		\small
		\begin{tabular}{p{0.25\linewidth}p{0.28\linewidth}p{0.35\linewidth}}
			\toprule
			Failure condition & What fails & Required response \\
			\midrule
			Missing predicate & The failure cannot be expressed & Redesign predicate vocabulary \\
			Inadequate template & The failure is visible but not repairable & Expand or revise edit library \\
			Rule conflict & Corrective rule is blocked & Revise priorities or conflict relation \\
			Stochastic instability & Repair does not survive held-out seeds & Increase replications, revise acceptance criteria \\
			Simulation-to-deployment shift & Simulation evidence does not transfer & Validate on deployment-domain data \\
			\bottomrule
		\end{tabular}
		\caption{Negative cases and structural limits of the verification protocol.}
		\label{tab:negative_cases}
	\end{table}
	
	These negative cases are not ancillary limitations. They define the boundary of the contribution. The framework supports bounded controller verification only when the relevant failure is expressible in the predicate vocabulary, diagnosable from logged controller behavior, correctable by the finite edit library, and stable under held-out evaluation.
	
\section{Experimental Results: Detectability, Repairability, and Rejection Cases}
\label{sec:inventory_results}

This section reports the results obtained with the inventory-control benchmark. The purpose of the experiments is not to show that the symbolic controller is optimal. The purpose is to test whether controller-level failures can be detected, mapped to finite edits, and evaluated on held-out stochastic seeds under declared diagnostic thresholds and guardrails.

Across the experiments, the repair protocol follows the same structure. First, the controller is evaluated on diagnostic seeds. Second, diagnostic failures are identified from the empirical evaluation quantile. Third, the coaching function selects one admissible edit from the finite edit library. Fourth, the original and revised controllers are compared on held-out seeds. A repair is accepted only if the target metric improves, the target threshold is restored, all guardrails remain within threshold, and the revised rule base remains syntactically valid.

\subsection{Constraint-Induced Failures}

The first two experiments tested the controller under replenishment envelopes that were too restrictive relative to expected demand. With eight inventory units and baseline demand near 20 units per item, expected aggregate demand is approximately 160 units per period. In the first experiment, replenishment capacity was set to 80 and the budget to 900. In the second experiment, these values were relaxed to 130 and 1500. Both settings remained insufficient to eliminate shortage and tracking failures.

In the first experiment, the diagnostic run detected stockout, service-loss, and tracking failures. The coaching function selected a stockout-increase rule:
\[
\texttt{IF any\_stockout\_risk THEN increase\_order}.
\]
The repair was rejected on held-out seeds. In the no-repair condition, the held-out evaluation statistics were
\[
SR = 0.9823,\qquad SL = 77.48,\qquad TE = 73.67.
\]
The coached repair did not restore any of these target metrics:
\[
SR = 0.9823,\qquad SL = 77.48,\qquad TE = 73.67.
\]
The conflict rate also increased from 0.1317 to 0.2307, although it remained below the declared conflict threshold.

In the second experiment, the replenishment envelope was relaxed. Failure severity decreased, but the same pattern remained. The no-repair condition produced
\[
SR = 0.9720,\qquad SL = 28.38,\qquad TE = 71.43.
\]
The stockout-increase repair again failed to restore the target thresholds:
\[
SR = 0.9730,\qquad SL = 28.38,\qquad TE = 71.43.
\]
The numeric controller and random-edit baselines reduced some metrics slightly, but they did not restore the declared stockout or tracking thresholds. These two experiments therefore provide negative evidence. They show that a failure may be observable through the diagnostic vocabulary while remaining non-repairable by one finite rule edit because the feasible replenishment envelope is insufficient.

\subsection{Damaged Stockout-Priority Rule}

The third experiment deliberately removed the stockout-priority rule:
\[
\texttt{r\_stockout\_priority}.
\]
This setting used the more permissive replenishment envelope, with capacity 160 and budget 1800. The diagnostic run detected stockout and tracking failures. The coaching function selected the same stockout-increase edit:
\[
\texttt{IF any\_stockout\_risk THEN increase\_order}.
\]

The repair produced mixed effects on held-out seeds. Service loss improved from
\[
SL = 0.2400
\]
to
\[
SL = 0.1360,
\]
and tracking error improved from
\[
TE = 52.71
\]
to
\[
TE = 50.99.
\]
However, stockout rate remained far above threshold and slightly worsened:
\[
SR = 0.8938 \rightarrow 0.8981.
\]
The conflict rate also increased from
\[
0.1571
\]
to
\[
0.2780,
\]
which exceeded the declared conflict threshold of 0.25. The repair was therefore rejected.

The original-rule restoration baseline restored \texttt{r\_stockout\_priority}. This reduced the conflict rate to 0.1074 and produced low service loss, but it did not restore the stockout or tracking thresholds:
\[
SR = 0.8724,\qquad TE = 51.73.
\]
This result indicates that, under the tested setting, \texttt{r\_stockout\_priority} was not the sole or sufficient cause of the observed stockout and tracking failures.

\subsection{Damaged Smoothing Rule}

The fourth and fifth experiments removed the smoothing rule:
\[
\texttt{r\_smooth}.
\]
This created a clearer rule-level damage condition. Without the smoothing rule, held-out order volatility was high:
\[
OV = 20.62,
\]
above the declared threshold
\[
\theta_{OV}=12.
\]

In the fourth experiment, the original coaching priority was retained. Because stockout and tracking failures were also present, the coaching function selected the stockout-increase edit rather than a smoothing edit. This was the wrong local repair target for the volatility failure. The coached repair reduced stockout rate from 0.7852 to 0.7473 and slightly reduced tracking error from 47.08 to 46.53, but it did not repair order volatility:
\[
OV = 20.62 \rightarrow 21.36.
\]
The repair was rejected.

The original-rule restoration baseline restored \texttt{r\_smooth}. This reduced held-out order volatility to
\[
OV = 11.46,
\]
which is below the declared volatility threshold. This indicates, under the declared simulator intervention, that the removed smoothing rule was relevant to the observed volatility failure. The claim is limited to the tested rule base, stochastic seeds, thresholds, and inventory-control configuration; it is not a general causal claim about inventory systems.

In the fifth experiment, the coaching priority was revised so that volatility failures select a smoothing edit before stockout-increase edits. The coaching function selected:
\[
\texttt{add\_r\_repair\_volatility\_smooth}.
\]
This edit added the rule:
\[
\texttt{IF any\_order\_volatility\_high THEN smooth\_order}.
\]
The coached repair produced the same volatility value as the original-rule restoration baseline:
\[
OV = 11.46.
\]
Thus, the protocol locally repaired the volatility failure by one finite edit.

However, the full controller was still not globally accepted. Stockout rate and tracking error remained above their thresholds:
\[
SR = 0.8724,\qquad TE = 51.73.
\]
The correct interpretation is therefore local rather than global: the volatility failure was one-step repairable, but the controller as a whole remained non-working under the joint acceptance predicate.

Table~\ref{tab:inventory_volatility_repair} reports the main held-out evaluation statistics for the fifth experiment. The table uses the empirical evaluation quantile reported by the simulation code. It shows that the smoothing edit restored the order-volatility threshold and preserved the conflict guardrail, but did not restore the stockout-rate or tracking-error thresholds.

\begin{table}[t]
	\centering
	\small
	\caption{Held-out evaluation statistics for the smoothing-rule damage experiment with revised coaching priority.}
	\label{tab:inventory_volatility_repair}
	\begin{tabular}{lrrrr}
		\toprule
		Metric & No repair & Coached repair & Threshold & Outcome \\
		\midrule
		Stockout rate & 0.7852 & 0.8724 & 0.1000 & Failed \\
		Tracking error & 47.08 & 51.73 & 30.00 & Failed \\
		Order volatility & 20.62 & 11.46 & 12.00 & Repaired \\
		Service loss & 0.0855 & 0.3948 & 5.00 & Within threshold \\
		Conflict rate & 0.1333 & 0.1074 & 0.2500 & Within threshold \\
		\bottomrule
	\end{tabular}
\end{table}

\subsection{Summary of Experimental Outcomes}

Table~\ref{tab:inventory_experiment_summary} summarizes the five rerun experiments. The fifth experiment is the strongest positive case, but only under a local interpretation. It shows that a specific rule-level failure can be repaired by a finite admissible edit when the failure is represented in the predicate vocabulary and the edit library contains an appropriate smoothing rule. The smoothing failure was detected, the coaching function selected the appropriate smoothing edit, and the held-out volatility metric was restored below threshold. At the same time, the experiment shows why the framework must distinguish local repair from global controller acceptance. Repairing volatility did not repair stockout and tracking failures.

\begin{table}[h]
	\centering
	\small
	\setlength{\tabcolsep}{3pt}
	\renewcommand{\arraystretch}{1.15}
	\caption{Summary of inventory-control repair experiments based on rerun outputs.}
	\label{tab:inventory_experiment_summary}
	\begin{tabular}{p{0.15\linewidth}p{0.22\linewidth}p{0.18\linewidth}p{0.25\linewidth}p{0.13\linewidth}}
		\toprule
		Experiment & Condition & Selected edit & Local result & Global result \\
		\midrule
		1 & \makecell[l]{Restrictive\\envelope} & Stockout increase & No target repair & Rejected \\
		2 & \makecell[l]{Relaxed\\envelope} & Stockout increase & No target repair & Rejected \\
		3 & \makecell[l]{Remove\\\texttt{r\_stockout\_priority}} & Stockout increase & Partial metric improvement & Rejected \\
		4 & \makecell[l]{Remove \texttt{r\_smooth}\\stockout-first coach} & Stockout increase & Wrong repair target & Rejected \\
		5 & \makecell[l]{Remove \texttt{r\_smooth}\\volatility-first coach} & Smoothing rule & Volatility repaired & Rejected \\
		\bottomrule
	\end{tabular}
\end{table}

\subsection{Interpretation}

The experimental results support three conclusions.

First, detectability is weaker than repairability. Stockout, service-loss, tracking, and volatility failures were detected by the diagnostic metrics, but not all detected failures were repairable by one admissible edit.

Second, repairability depends on the source of failure and on the coaching selector. When the failure was induced by an insufficient replenishment envelope, rule edits did not restore the target metrics. When the smoothing rule was removed, restoring or adding a smoothing rule repaired the volatility metric. However, this local repair was obtained only after the coaching priority was changed to select volatility repair before stockout repair.

Third, held-out evaluation prevents overclaiming. Several edits produced directional improvements, but were rejected because they failed to restore thresholds or because they worsened guardrail metrics. This is the intended behavior of the protocol. The contribution is not an assertion that symbolic coaching always repairs adaptive controllers. The contribution is a bounded procedure for identifying which failures are observable, which are locally repairable, and which remain outside the declared repair capacity of the finite edit library.
	
	\section{Scope of Claims}
	
	The experiments support a limited interpretation. If a controller failure is expressed in the diagnostic vocabulary, if the relevant condition is visible in the explanation logs, if the edit library contains an appropriate correction, and if the coaching selector chooses that correction, then the failure may be one-step repairable. The smoothing-rule damage experiment provides one positive case. After \texttt{r\_smooth} was removed, order volatility exceeded its declared threshold. When the coaching selector prioritized volatility repair, a finite smoothing edit restored the held-out order-volatility metric below threshold. This supports the claim that some controller-level failures can be locally addressed without unrestricted human-in-the-loop judgment.
	
	The experiments do not support stronger claims. They do not show that all agentic failures are detectable. They do not show that all detected failures are repairable. They do not show that an available repair template will be selected under every coaching priority. They do not show that local repair implies global controller acceptance. They do not show that simulation results transfer directly to production. They do not prove that finite symbolic rules are sufficient for general agentic AI safety.
	
	The negative and rejected cases are part of the result. Resource-induced failures remained non-repairable by one rule edit. Removing \texttt{r\_stockout\_priority} produced failures that were not fully restored by either the coached stockout-increase edit or original-rule restoration. Removing \texttt{r\_smooth} produced a volatility failure that was repairable only when the coaching selector selected the smoothing edit. Even then, the controller was globally rejected because stockout and tracking failures remained above threshold. The supported claim is therefore narrower: a bounded class of controller failures can be made observable, locally revisable, and testable under controlled conditions, but local repair does not imply complete controller repair.

	\section{Conclusion}
	
	This paper formulates a simulation-based verification protocol for adaptive agentic controllers represented through finite rules, diagnostic predicates, explanation logs, and held-out re-evaluation. The central claim is bounded: some controller-level failures can be detected, locally repaired, or rejected without unrestricted human-in-the-loop judgment when the relevant failure is expressible in the predicate vocabulary, traceable in the explanation log, and correctable through a finite edit library.
	
	For practitioner use, the framework should be read as a qualification layer rather than as an autonomy guarantee. It helps structure questions that enterprise users must ask before deployment: what is observable, what is logged, what is bounded by guardrails, what can be revised without free-form human intervention, what survives held-out testing, and what remains outside the verification envelope. This makes the method relevant to agentic-AI adoption programs, but only as one component of a broader operating model that also includes security, legal review, data governance, monitoring, and production integration.
	
	The inventory-control experiments show that this distinction matters. Resource-induced stockout and tracking failures were detectable but not one-step repairable under the tested envelopes. A stockout-increase edit produced partial effects but was rejected when thresholds or guardrails were not restored. By contrast, the removal of \texttt{r\_smooth} created an order-volatility failure that was locally repaired by adding a smoothing rule; the same repair did not globally accept the controller because stockout and tracking failures remained. The framework therefore reframes part of the agentic AI deployment problem as a controller-level verification problem. Instead of asking whether an agentic prototype is generally trustworthy, it asks which failures are observable, which are attributable to controller logic, which are repairable by finite edits, and which remain outside the architecture. This provides a disciplined basis for studying verification, contestability, and bounded repair in adaptive agentic systems.
	
	\section*{Competing Interests}

The author declares no competing interests.

\section*{AI-Use Disclosure}

AI-based writing assistance has been used for language editing, structuring, and drafting support. All scientific claims, formal definitions, methodological choices, and final responsibility remain with the author.
	
	\section*{Data and Materials Availability}
	
	The preprint is intended to be accompanied by a public repository containing the executable simulation code, rule-base definitions, diagnostic-threshold configuration files, seed lists, parameter files, table-generation scripts, and figure-generation scripts.

		\section*{Reproducibility Specification}
	\label{sec:reproducibility}
	
	The experimental results are intended to be reproducible from the released notebooks and exported output files. Each experiment is implemented as a self-contained Jupyter notebook containing the simulator, controller, diagnostic functions, finite edit library, coaching function, baseline comparison code, and export routines. The rerun package contains five notebooks and their corresponding JSON and CSV outputs:
	\[
	\begin{aligned}
		&\texttt{controller\_change\_assurance.ipynb},\\
		&\texttt{inventory\_verification\_self\_contained.ipynb},\\
		&\texttt{inventory\_verification\_self\_contained-2ndExperiment.ipynb},\\
		&\texttt{inventory\_verification\_self\_contained-3ndExperiment.ipynb},\\
		&\texttt{inventory\_verification\_self\_contained-4thExperiment.ipynb},\\
		&\texttt{inventory\_verification\_self\_contained-5thExperiment.ipynb}.
	\end{aligned}
	\]
	Each notebook exports a repair-report JSON file and a baseline-metrics CSV file. The JSON file records the selected edit, diagnostic failures, target metrics, acceptance status, before/after held-out summaries, and rule-consistency status. The CSV file records the held-out evaluation metrics for the no-repair, coached-repair, random-edit, manual-tuning-proxy, numeric-controller, and, where applicable, original-rule-restoration baselines.
	
	The experiments use the same core simulator and differ only in replenishment envelope, rule damage, and coaching priority. Across experiments, the number of inventory units is
	\[
	N=8,
	\]
	the horizon is
	\[
	T=120,
	\]
	initial inventory is 70 per unit, baseline demand is 20 per unit, demand standard deviation is 5, unit cost is 10, the lower inventory threshold is 35, the upper threshold is 110, and the target inventory is 75. Diagnostic seeds are
	\[
	\{1,2,3,4,5\},
	\]
	held-out evaluation seeds are
	\[
	\{101,\ldots,120\},
	\]
	and the stochastic tolerance is
	\[
	\alpha = 0.05.
	\]
	The reported evaluation statistic is the empirical $(1-\alpha)$ quantile, denoted \texttt{q\_eval} in the exported files.
	
	Table~\ref{tab:inventory_reproducibility} lists the minimum artifacts required to reproduce the reported experiments.
	
	\begin{table}[!htbp]
		\centering
		\small
		\begin{tabular}{p{0.30\linewidth}p{0.60\linewidth}}
			\toprule
			Artifact & Required content \\
			\midrule
			
			Simulation notebooks
			&
			Five self-contained Jupyter notebooks implementing the inventory-control simulator, stochastic demand process, capacity and budget constraints, finite symbolic controller, predicate abstraction, coaching function, diagnostics, baseline comparisons, and export routines. \\
			
			Experiment configurations
			&
			Declared configuration for each experiment: restrictive envelope $(K=80,B=900)$; relaxed envelope $(K=130,B=1500)$; removal of \texttt{r\_stockout\_priority} under $(K=160,B=1800)$; removal of \texttt{r\_smooth} with stockout-first coaching under $(K=160,B=1800)$; and removal of \texttt{r\_smooth} with volatility-first coaching under $(K=160,B=1800)$. \\
			
			Controller specification
			&
			Initial rule base $R$, damaged rule base where applicable, predicate vocabulary $\mathcal{P}$, action vocabulary $\mathcal{A}$, conflict relation $\perp$, priority ordering $\pi$, rule-resolution procedure, finite edit library, and coaching-priority rule used in each notebook. \\
			
			Randomness
			&
			Diagnostic seed list $\{1,\ldots,5\}$, held-out seed list $\{101,\ldots,120\}$, random seed for the random-edit baseline, number of stochastic replications, and the random-number generator used by the notebook runtime. \\
			
			Model parameters
			&
			Number of inventory units $N=8$, horizon $T=120$, initial inventories $I_{i,0}=70$, baseline demands $\bar{D}_i=20$, demand standard deviation 5, lower threshold $L_i=35$, upper threshold $U_i=110$, target inventory $I_i^\ast=75$, unit costs $c_i=10$, replenishment capacity $K$, budget $B$, and allocation parameters. \\
			
			Diagnostic thresholds
			&
			Declared thresholds for stockout rate, overstock rate, service loss, working-capital excess, tracking error, order volatility, budget-violation rate, capacity-violation rate, conflict rate, stochastic tolerance $\alpha=0.05$, and the empirical evaluation quantile rule. \\
			
			Baselines
			&
			No-repair, coached-repair, random-edit, manual-tuning-proxy, numeric-controller, and, for damaged-rule experiments, original-rule-restoration baselines. Each baseline must use the same held-out seed list and diagnostic thresholds. \\
			
			Exported outputs
			&
			For each notebook, a repair-report JSON file and a baseline-metrics CSV file. The JSON file records selected edit, target metrics, diagnostic failures, acceptance status, rule consistency, and before/after summaries. The CSV file records held-out \texttt{mean}, \texttt{sd}, \texttt{q\_eval}, \texttt{q95}, and applied edit for each variant and metric. \\
			
			Environment
			&
			Python version, Jupyter runtime, package versions for \texttt{numpy}, \texttt{pandas}, and plotting libraries if figures are regenerated, operating-system note, and repository or archive version for the submitted rerun package. \\
			
			\bottomrule
		\end{tabular}
		\caption{Minimum reproducibility specification for the five inventory-control rerun experiments.}
		\label{tab:inventory_reproducibility}
	\end{table}
	
	Exact numerical equality may depend on software versions and random-number implementation, but the qualitative claims in the paper require the following reproducibility checks. First, the restrictive and relaxed resource-envelope experiments must diagnose stockout, service-loss, and tracking failures and reject the stockout-increase repair. Second, the \texttt{r\_stockout\_priority} damage experiment must reject the stockout-increase repair because stockout and tracking thresholds are not restored and the conflict guardrail is violated. Third, the stockout-first \texttt{r\_smooth} damage experiment must show that the available smoothing repair is not selected, while original-rule restoration reduces order volatility below threshold. Fourth, the volatility-first \texttt{r\_smooth} damage experiment must select the smoothing edit and reduce held-out order volatility from above threshold to below threshold, while still rejecting the controller globally because stockout and tracking failures remain.
	
	\newpage
	\bibliographystyle{unsrtnat}

\end{document}